\documentclass{article}
\usepackage{iclr2027_conference,times}

\usepackage{graphicx}
\usepackage{booktabs}
\usepackage{float}
\usepackage{amsmath,amssymb}
\usepackage{array}
\usepackage{microtype}
\usepackage{url}
\usepackage{hyperref}
\usepackage{xcolor}
\hypersetup{colorlinks=true,allcolors=black}

\title{Right Order, Wrong Scale:\\Auditing LLM Judges\\for Occupational AI Measurement}

\author{Harry Lyu\\
Massachusetts Institute of Technology\\
\texttt{hlyu@mit.edu}
\And
Neil Thompson\\
Massachusetts Institute of Technology\\
\texttt{neil\_t@mit.edu}
}

\iclrfinalcopy 

\newcommand{\pacc}{\mathrm{PairAcc}_{1/2}}
\newcommand{\dataset}{\textsc{O*NET-Bench}}

\begin{document}
\maketitle
\lhead{Preprint. Under review.}

\begin{abstract}
LLM judges are increasingly used to assess whether AI outputs meet workplace requirements, but agreement on response rankings does not establish agreement on acceptance rates or occupational aggregates. We introduce O*NET-BENCH, an audit suite derived from an existing survey of 45,796 worker ratings, and evaluate 33 pre-existing judge configurations across six model families on 4,501 test ratings. Twenty-five configurations achieve tie-aware pair accuracy of at least 0.60, although a train-fitted response-only TF--IDF baseline nearly matches the strongest judge. Despite this ordering agreement, judges estimate that 3.0\%--97.9\% of responses are acceptable, compared with 61.1\% for occupation-matched workers. In one fine-tuned lineage, changing from pointwise scoring to a bundled few-shot/listwise protocol improves response ordering while reducing agreement with worker means at the task and occupation levels; this reversal replicates on a task- and worker-disjoint validation split under prespecified criteria. Cross-validated calibration largely removes mean bias, but calibrated scores explain at most 8.5\% of individual worker-rating variance. Prediction-assisted estimation yields at most small precision gains at the studied label budgets. These results show that ranking agreement alone is insufficient for occupational measurement. Judges should be validated against the acceptance rates and aggregates their scores will be used to estimate.
\end{abstract}

\section{Introduction}

How should evaluations determine whether AI outputs meet workplace requirements? Occupational and professional benchmarks assess model outputs on work-related tasks \citep{openai2025gdpval,wang2026profbench,shao2025workbank,li2026jobbench}, while exposure studies and usage analyses address related questions about AI and work \citep{eloundou2023gpts,felten2023ai,gan2026execution,anthropic2025econindex}. These approaches measure different quantities. We focus on evaluations that use numerical judgments to estimate the share of responses that meet an acceptance threshold or to summarize performance across tasks and occupations. Interpreting such estimates requires validating the scores from which they are calculated.

These scores increasingly come from an LLM judge, because expert grading is a major cost in many such evaluations. Judges are commonly selected by agreement with human ratings on individual responses, through correlation, pairwise accuracy, or recovered rankings. Ranking agreement establishes whether a judge prefers the same responses as workers. It does not establish whether the judge assigns comparable scores or identifies the same responses as acceptable. We call judge selection that evaluates ordering without checking these downstream quantities \emph{scale-blind selection}. This paper audits its consequences.

\begin{figure}[t]
\centering
\includegraphics[width=\textwidth]{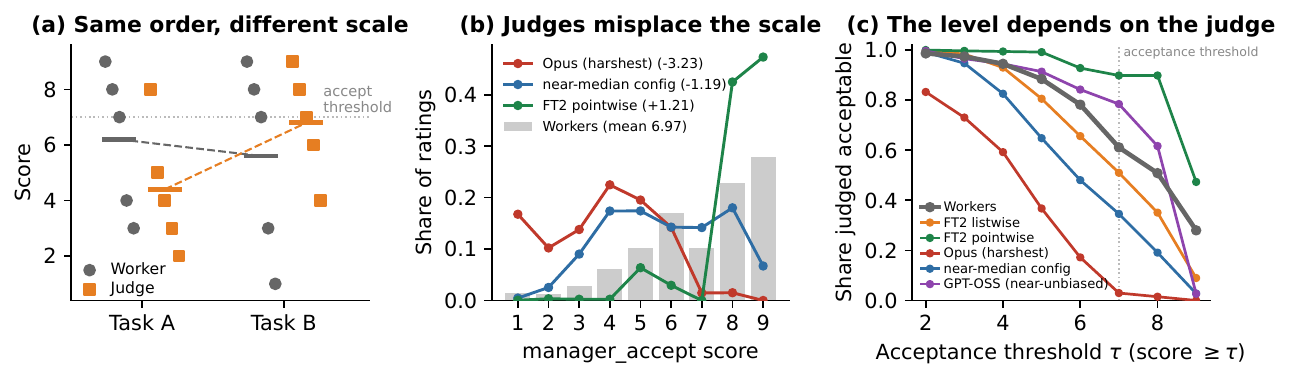}
\caption{The scale problem. (a)~A real pair of test-split task instances, responses sorted by worker score: the judge reproduces every within-task ordering exactly, yet the task means (horizontal markers, joined by dashed lines) move oppositely and Task~A's majority verdict flips. (b)~Test-split score distributions: judges reproduce neither the location nor the shape of the worker distribution; parenthesized values are mean bias. (c)~The quantity these evaluations report, the share judged acceptable at threshold $\tau$, depends on the grader; judge curves are steep near the acceptance threshold, so small location errors become large level errors.}
\label{fig:scale}
\end{figure}

Consider a worker with paralegal experience who rates five AI-drafted case summaries 5, 6, 7, 8, and 9 on a nine-point acceptability scale where a score of 7 or higher counts as acceptable, and a judge that rates the same five summaries 1, 2, 3, 4, and 5. By any purely rank-based selection metric the judge is perfect: it reproduces the worker's ordering exactly. But the worker says three of the five summaries are usable as-is; the judge says none is. Aggregated over an occupation, the two instruments yield different conclusions from the same responses, and ranking agreement cannot distinguish them, because within-task order is invariant to shifts and stretches of the scale. Figure~\ref{fig:scale}a shows a real test-split pair with this structure (civil engineering and psychology teaching, ten-exemplar fine-tuned listwise judge); over 7{,}000 such pairs occur across the 32 near-complete configurations (pairs where one judge reproduces both instances' worker-decisive orderings while the instance means move oppositely by $\geq$0.6 points). The example separates \emph{ordinal} validity (which response is better), \emph{cardinal} validity (how good each response is, in location and spread), and \emph{aggregate} validity (whether task and occupation means agree).

To study these forms of agreement in occupational evaluation, we construct \dataset{} from the existing worker-evaluation corpus of \citet{mertens2026crashing}. The source contains 45{,}796 ratings of AI responses to O*NET-derived workplace tasks \citep{onet2024}, provided by workers with self-reported experience in the corresponding occupations. Our audit suite aligns responses with ratings, defines task- and worker-disjoint splits, and supports evaluation from individual responses to job families. On the test split, we compare 33 pre-existing judge configurations spanning six model families and six train-fitted controls.

The audited judges contain information about response ordering, although a response-only TF--IDF baseline nearly matches the strongest configuration. Their numerical scores agree less well with the worker reference: mean-bias intervals exclude zero for all 33 configurations, and estimated acceptance shares vary substantially. Ranking performance is only weakly associated with these differences in score placement. In one fine-tuned lineage, a bundled few-shot/listwise protocol improves response-level ordering while reducing agreement with worker means at the task and occupation levels. This reversal replicates under prespecified criteria on the validation split; a separate, non-preregistered eight-arm check reproduces the main severity pattern. Calibration largely removes mean bias, but the calibrated scores explain at most 8.5\% of individual worker-rating variance.

We contribute \dataset{}, an audit suite derived from an existing worker-evaluation corpus (no new human ratings were collected); a comparison of 33 judge configurations across ordering, score levels, and aggregates, including a prespecified reversal replication in one fine-tuned lineage; and an evaluation of calibration and prediction-assisted estimation. Appendix~\ref{app:standard} translates the findings into a six-item reporting standard.

\section{Related work}

\paragraph{Judge selection and evaluation protocols.}
Work on MT-Bench, Chatbot Arena, G-Eval, and RewardBench has developed influential approaches to model evaluation \citep{zheng2023judging,chiang2024arena,liu2023geval,lambert2025rewardbench}, while Prometheus~2 and JudgeLM train specialized evaluators \citep{kim2024prometheus,zhu2025judgelm}; our audit covers general foundation models adapted as judges, not purpose-built evaluators of this kind. Listwise protocols place candidate responses in a shared context \citep{yang2024linkage}. Evaluation criteria in this literature include correlations, preference accuracy, leaderboard agreement, and exact label agreement; ProfBench also uses macro-F1 and a bias index \citep{wang2026profbench}. These criteria provide different kinds of evidence. Our question is whether the evidence used to evaluate a judge supports the acceptance rates and occupational aggregates subsequently computed from its scores. Paired-comparison scaling in the style of Bradley--Terry identifies utilities only up to location and scale, so it does not by itself yield the absolute threshold quantities studied here.

\paragraph{Judge bias, severity, and calibration.}
Position, verbosity, self-preference, format, model-family, and leniency effects have been documented in judge evaluations \citep{shi2025position,dubois2024length,panickssery2024self,bavaresco2025judgebench}. Proposed responses address different problems: pairwise evaluation changes the elicitation format, whereas offset correction and recalibration adjust aspects of the numerical scale. Their effectiveness must nevertheless be assessed against the intended quantity on held-out data. We examine whether common judge configurations and calibration procedures support agreement with workers on response ordering, score levels, acceptance shares, and occupational aggregates.

\paragraph{Validation after aggregation and limits of substitution.}
JuStRank shows that instance-level judge quality and aggregate system ranking can diverge \citep{gera2025justrank}; rating-indeterminacy and human-uncertainty analyses complicate single gold labels \citep{guerdan2025indeterminacy,elangovan2025beyond}; agreement-metric guidance urges explicit scales, ties, and coverage \citep{rao2026agreement}; and judge identity itself changes the measurement \citep{yang2026measurement}. \citet{dorner2025limits} prove that a judge no more accurate than the evaluated model cannot reduce annotation requirements by more than half. We supply the empirical complement: the level error itself, its occupational consequences, and the small effective-sample value of a judge under prediction-powered inference \citep{angelopoulos2023ppi}.

\paragraph{Measurement across levels.}
Evaluation is measurement: evidence is valid for an interpretation and use, not for a score in the abstract \citep{wallach2025measurement}. Ordinal and interval information are distinct; reliability and invariance can differ between within-group and between-group components of the same data \citep{jak2017multilevel}; ecological correlations need not reproduce individual relationships \citep{robinson1950ecological}; and classical test theory supplies attenuation machinery for noisy targets \citep{spearman1904attenuation,brown1910spearman,cronbach1951alpha,shrout1979icc}. Our ordinal/cardinal/aggregate distinction is the judge-evaluation instance of this classical structure.

\paragraph{Occupational evaluation of AI.}
GDPval, the Anthropic Economic Index, exposure studies, ProfBench, WORKBank, and JobBench make occupational measurement consequential \citep{openai2025gdpval,anthropic2025econindex,eloundou2023gpts,gan2026execution,wang2026profbench,shao2025workbank,li2026jobbench}. We are not aware of a systematic test of whether automated judges reproduce occupation-matched workers' score levels across judge designs. That is the audit we perform, on a testbed whose target quantities come from an economic study of labor-market tasks \citep{mertens2026crashing}.

\section{Data and experimental design}
\label{sec:data}

\paragraph{Provenance.}
The ratings were collected by \citet{mertens2026crashing} in a large-scale survey studying AI automation through worker evaluations of labor-market tasks. No new human data were collected for this audit. Participants were recruited in the United States through Prolific, selected an occupation in which they reported experience, and confirmed the offered job matched that experience; submissions passed the source study's preregistered quality and experience screens \citep{mertens2026crashing}. The audit uses a frozen extraction provided by the source study's authors (45{,}796 ratings, 47 generator variants; snapshot of 2026-08-18). The source publication analyzes an earlier snapshot of the same ongoing collection (17{,}205 post-screen ratings), so its published counts are smaller. The workers define the audit's estimand: we treat them as \emph{target raters} with self-reported relevant experience and measure agreement with them, not with an assumed objective truth. For the audit, \dataset{} adds per-response alignment of ratings with AI outputs, disjoint splits, fixed prompts and an evaluation protocol over the aggregation hierarchy, and a prediction matrix covering every configuration.

\paragraph{Hierarchy and target.}
Each observation links an O*NET occupation and task, a concrete task instance, an AI-generated response, and one worker rating. The aggregation hierarchy is response $\to$ task instance $\to$ task ID $\to$ occupation $\to$ job family; each instance contains approximately five responses. The primary target, \texttt{manager\_accept}, uses a nine-point scale: 1 indicates that the response requires complete rework, 7 that it is minimally sufficient without edits, and 9 that it is of superior quality without edits. Following the source study, we set the acceptance threshold to $\tau=7$ and classify scores of 7 or higher as acceptable. The training, validation, and test splits contain 36{,}649, 4{,}646, and 4{,}501 ratings, respectively. Task IDs and workers are disjoint across splits, while occupations and generator variants overlap (Table~\ref{tab:splits}). The design therefore evaluates new tasks and workers within a shared occupation and generator universe. Validation loss was monitored during fine-tuning but was not used to select checkpoints. Subsequent validation analyses include the prespecified reversal replication and separately identified exploratory checks.

\paragraph{Judge configurations.}
We audit stored predictions from 33 pre-existing configurations spanning six model families: Claude (Haiku 4.5 and Opus 4.7), GPT-5.4-mini, GPT-OSS-120B, DeepSeek-V4-Pro, Llama-3.3-70B, and Qwen3.5-9B, including one- and two-epoch LoRA checkpoints \citep{hu2022lora}. Variations include exemplars, pointwise or listwise presentation, rubric and other prompting scaffolds, reasoning and decoding settings, self-consistency, ensembling, and fine-tuning. The configurations are related experimental conditions, not 33 independent replications.

Thirty-two configurations cover at least 99.8\% of test rows; we call these \emph{near-complete}. One reasoning variant covers 22.3\% because of the serving issue documented in Appendix~\ref{app:serving}. Unless otherwise stated, configuration summaries include all 33 arms, evaluating the partial-coverage arm on its available rows; cross-configuration comparisons use the 32 near-complete arms. Six train-fitted controls (training-set mean, response-position means, occupation means, length-plus-position regression, and response-only and full-context TF--IDF ridge regressions) measure agreement obtainable from simple statistical structure. Appendix~\ref{app:registry} reports the full registry.

Pointwise prompts score one response at a time and elicit the survey's multi-field schema. Listwise prompts score the responses of a task instance (typically five) together, in a fixed, non-randomized order (Appendix~\ref{app:mechanism}), and elicit only \texttt{manager\_accept} (abridged templates: Appendix~\ref{app:prompts}). A base-model $2{\times}2$ comparison varies presentation and exemplars; because every listwise prompt elicits only \texttt{manager\_accept}, its presentation factor also bundles the schema change. The primary fine-tuned contrast changes presentation, exemplars, and elicitation schema, so we interpret it as a bundled protocol change rather than an isolated effect of listwise presentation.

\section{Validity must be measured at the intended level}
\label{sec:metrics}

Let $y_{ti}$ and $\hat y_{ti}$ be worker and judge scores for response $i$ in task instance $t$. Table~\ref{tab:design} summarizes the audit design. Each metric below matches one way judge scores are consumed downstream.

\begin{table}[!ht]
\centering
\footnotesize
\setlength{\tabcolsep}{3pt}
\caption{Audit overview. All test-split analyses are exploratory; preregistered validation covers the primary fine-tuned protocol contrast, in its response-ordering and aggregate-agreement directions (\S\ref{sec:reversal}), and the separate eight-arm severity check was not preregistered.}
\label{tab:design}
\begin{tabular}{@{}lll@{}}
\toprule
Component & Main measures & Evaluation level \\
\midrule
Ordering agreement & Tie-aware pair accuracy, coverage, $\rho$ & response; within-instance pairs \\
Score-level agreement & Mean bias, SD ratio, acceptance share & response \\
Aggregate agreement & Cell-mean corr., classification disagreement & task; occupation; family \\
Calibration and label efficiency & CV bias and $R^2$; PPI precision, coverage & response; pooled \\
\bottomrule
\end{tabular}
\end{table}

\paragraph{Ordinal validity.}
For every worker-decisive pair within an instance, the judge earns 1 for the same ordering, 0 for the reverse, and $\tfrac12$ for a tie:
\begin{equation}
\pacc=\frac{\sum_{t,i<j:y_{ti}\neq y_{tj}} \bigl[\mathbf{1}\{s_{tij}=\hat s_{tij}\}+\tfrac12\,\mathbf{1}\{\hat s_{tij}=0\}\bigr]}{\sum_t |\{(i,j): i<j,\, y_{ti}\neq y_{tj}\}|},
\end{equation}
where $s_{tij}=\mathrm{sign}(y_{ti}-y_{tj})$ and $\hat s_{tij}=\mathrm{sign}(\hat y_{ti}-\hat y_{tj})$. We report $\pacc$ together with conditional accuracy on judge-decisive pairs and \emph{pair coverage}, the share of decisive pairs on which the judge expresses an ordering. Response-level Spearman $\rho$ is the pooled rank correlation over all covered test rows. A constant judge cannot rank, and coverage makes that visible.

\paragraph{Cardinal validity.}
Signed mean bias, MAE, the judge/worker score-SD ratio, the full score distribution, and threshold-crossing rates $\Pr(\hat y \geq \tau)$ for $\tau\in\{5,\dots,8\}$. Cardinal validity is not mean bias alone: compressed distributions concentrate mass near the threshold, so a judge with a near-zero mean bias can still misreport the thresholded level badly (GPT-OSS below: bias $+0.12$, acceptance-share difference $+17.3$ percentage points). \emph{Cardinal} is operational shorthand for the numerical levels and spread consumed downstream.\footnote{Not a claim that the nine-point survey scale has interval properties.}

\paragraph{Aggregate validity.}
Worker and judge cell means at task ID, occupation, and job family (minimum 5, 5, and 8 ratings; 515, 307, and 21 evaluable cells, with one of the 22 job families below the minimum), compared by correlation, and by threshold-based occupation classification: the share of occupations in which a strict majority of sampled responses is rated acceptable.

\paragraph{Ordering does not identify level.}
For any scores $x_{ti}$ and any desired task means $m_t$, the transformation $x'_{ti}=a_t x_{ti}+b_t$ with $a_t>0$, $b_t=m_t-a_t\bar x_t$ preserves every within-task rank and pairwise decision while setting $\bar x'_t=m_t$. Occupation means, as weighted averages of task means, move freely as well (Figure~\ref{fig:scale}a). Perfect ordinal validity therefore guarantees nothing about cardinal or aggregate validity. This construction is stated for real-valued scores; restricting outputs to the nine-point scale limits the admissible transformations but does not remove the distinction, and Figure~\ref{fig:scale}a gives a bounded-scale example in which within-task ordering is preserved while task means and an acceptance verdict change. To locate the empirical differences, we also use one-way variance shares $V_G(z)=\sum_g n_g(\bar z_g-\bar z)^2/\sum_i(z_i-\bar z)^2$ for groupings $G$ (descriptive and non-additive).

\paragraph{Uncertainty and epistemic status.}
Response-level 95\% intervals use 1{,}000 bootstrap draws clustered by task ID; paired contrasts use common rows; occupation-classification intervals resample occupation cells; and calibration uses task-grouped cross-validation. Comparisons across the related judge configurations are descriptive. Each response has only one worker rating, so the audit measures agreement with an observed worker reference rather than an error-free target. Occupation-level split-half reliability is low (Spearman--Brown estimate 0.093; Appendix~\ref{app:reliability}), but this estimate does not quantify reliability at every aggregation level. Under classical assumptions, independent mean-zero target noise attenuates population correlations and contributes no expected mean bias; finite-sample uncertainty and nonlinear thresholding still require separate treatment. We therefore report clustered intervals and sensitivity analyses for acceptance shares and occupation classifications (\S\ref{sec:consequences}; Appendix~\ref{app:occ}). All test-split analyses are exploratory or robustness checks because these configurations were inspected repeatedly. Only analyses explicitly identified as preregistered had their directions and numerical thresholds fixed before evaluation on the validation split.

\section{Results}
\label{sec:results}

Sections~\ref{sec:ordering}--\ref{sec:decoupling} establish the ordering signal, the systematic departures in score placement, and the decoupling between the two. Sections~\ref{sec:interventions}--\ref{sec:calibration} evaluate prompting interventions and label-based repairs and trace the disagreement to the published acceptance share. Section~\ref{sec:reversal} presents the preregistered protocol reversal.

\subsection{Ordering signal is widespread but shallow}
\label{sec:ordering}

Ordering agreement is above chance for all 33 configurations: every task-clustered interval for tie-aware pair accuracy excludes 0.5. Twenty-five configurations reach at least 0.60, the median is 0.631, and the strongest near-complete configuration, the fine-tuned few-shot/listwise judge, reaches 0.658 with 0.912 pair coverage. Coverage exposes a limitation of conditional accuracy: Llama-3.3-70B orders only 5.1\% of worker-decisive pairs, so its conditional accuracy of 0.631 corresponds to a tie-aware accuracy of 0.507. Across six matched model comparisons, the few-shot/listwise protocol improves response-level Spearman correlation by 0.043--0.110 relative to the pointwise protocol, with every interval excluding zero (Appendix~\ref{app:pairs}). The primary fine-tuned contrast also improves response correlation on validation, by 0.033 $[0.003, 0.062]$ (Appendix~\ref{app:replication}). These comparisons describe the evaluated protocol changes; they do not all isolate presentation alone.

The signal is nonetheless shallow. A ridge regression on response-only TF--IDF features, fully supervised on the 36{,}649 training ratings (unlike the zero- and few-shot judges) and never shown an occupation label, reaches 0.643 pair accuracy at full pair coverage and 0.229 response Spearman against the top judge's 0.658 and 0.263; because generator variants overlap across splits, lexical features may also partly track generator identity (Appendix~\ref{app:mechanism}). Most configurations contain useful information about which response workers prefer, and the train-fitted bag-of-words control exceeds the median configuration while coming within 0.015 pair accuracy of the strongest.

\subsection{Score placement departs systematically from the target ratings}
\label{sec:severity}

\begin{figure}[t]
\centering
\includegraphics[width=\textwidth]{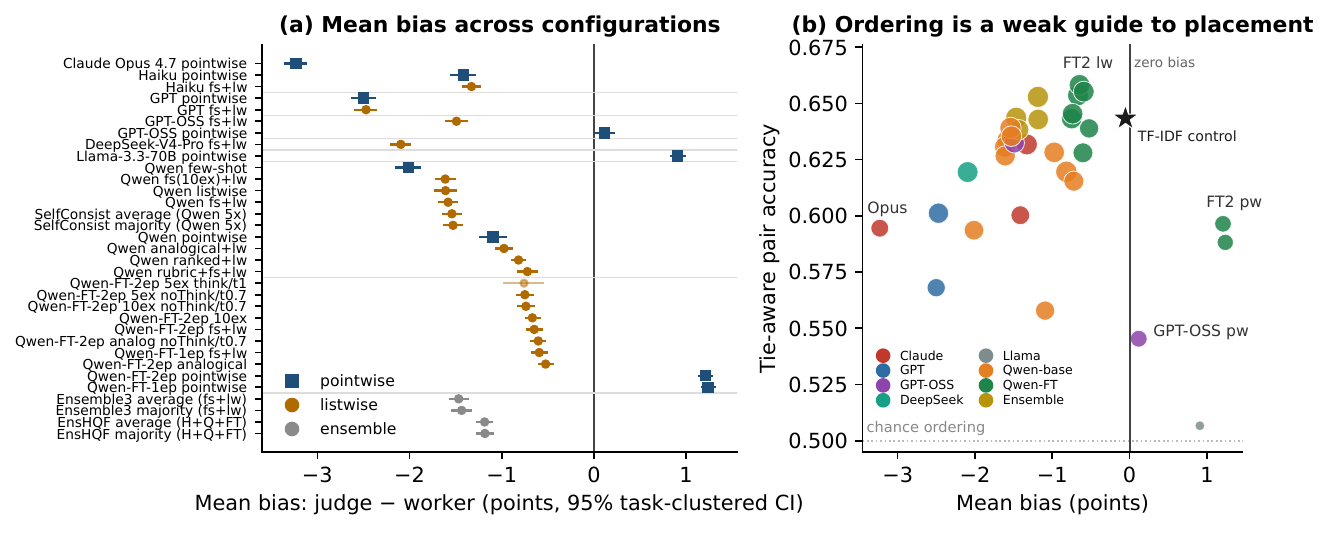}
\caption{Cardinal validity across the audit. (a)~Mean bias (judge $-$ worker) for all 33 configurations, with 95\% task-clustered intervals, grouped by model family. Squares are pointwise arms, and the faded row is the partial-coverage reasoning variant. Every interval excludes zero. (b)~Tie-aware pair accuracy against mean bias (point size: pair coverage).}
\label{fig:severity}
\end{figure}

Mean scores differ systematically from the worker reference (Figure~\ref{fig:severity}a). All 33 task-clustered mean-bias intervals exclude zero: 29 configurations score below workers on average and four score above them. The median bias is $-1.18$ rating points on the nine-point scale; 29 configurations are at least 0.5 rating points below workers and ten are at least 1.5 rating points below. Claude Opus 4.7 has the largest negative bias, $-3.23$ $[-3.36,-3.11]$, equivalent to 1.7 worker standard deviations, and assigns a lower score than the worker on 86\% of responses. The two fine-tuned pointwise checkpoints have positive biases of $+1.21$ and $+1.24$. Llama-3.3-70B has bias $+0.91$ but produces near-constant scores, whereas GPT-OSS-120B pointwise has the smallest absolute mean bias, $+0.12$ $[0.01, 0.22]$. Table~\ref{tab:registry} gives every configuration; Appendix~\ref{app:ft} describes the fine-tuned score concentration without identifying its causal mechanism.

GPT-OSS also shows why location is not the whole of cardinal validity. Its distribution is compressed (score-SD ratio 0.83), concentrating mass near the threshold, so a $+0.12$ mean bias becomes a 78.4\% acceptance rate against the workers' 61.1\%, a 17.3-percentage-point difference from the worker acceptance share for a judge with near-zero mean bias. Because listwise scores carry protocol-dependent location (\S\ref{sec:reversal}), we base the severity claim on the pointwise arms and read listwise biases as properties of the protocol.

Independent mean-zero rating noise contributes no expected mean difference (finite-sample variation aside), and a non-preregistered eight-arm validation check reproduces the main severity pattern: Haiku has bias $-1.41$ on both splits, GPT-5.4-mini has $-2.50$ on test and $-2.49$ on validation, and the fine-tuned pointwise-to-few-shot/listwise sign change of $+1.21$ to $-0.65$ reappears as $+1.26$ to $-0.70$ (Appendix~\ref{app:replication}). These results support the persistence of disagreement with the worker reference across splits.

\subsection{Ordering performance carries little information about placement}
\label{sec:decoupling}

Across the 32 near-complete configurations, pair accuracy is only weakly associated with signed bias (Pearson $-0.22$) and its magnitude (Spearman $-0.29$), and the individual contrasts are large (Figure~\ref{fig:severity}b): Opus and the pointwise fine-tune sit 4.4 rating points apart in mean bias at nearly identical pair accuracy (0.595 versus 0.596), and the judge with the smallest mean bias, GPT-OSS, is among the worst rankers. The same holds one level up: tie-aware pair accuracy tracks response-level Spearman at $\rho=0.91$ but shows little association with task-level ($-0.06$) or occupation-level agreement (0.00).

An exploratory selection experiment illustrates the practical consequence within this judge roster. Across 1{,}000 random half-splits of test tasks, we select a configuration on one half and evaluate it on the other (Appendix~\ref{app:selection}). Selection by pair accuracy yields a mean absolute acceptance-share error of 10.5 percentage points, compared with 5.8 when selecting directly for acceptance-share error; the respective 2.5th--97.5th percentile ranges across splits are $[7.7, 24.1]$ and $[3.6, 8.3]$. Direct selection reduces acceptance-share error by 4.7 percentage points on average, with a 0.018 reduction in pair accuracy. Selection by absolute mean bias always chooses GPT-OSS pointwise and yields a larger acceptance-share error of 17.3 percentage points. Because the configurations were previously inspected on the test split, these results describe selection within the audited roster rather than prospective performance on a new benchmark.

\subsection{Evaluated interventions do not consistently recover the worker scale}
\label{sec:interventions}

Across the evaluated groups, no intervention consistently recovers the worker scale (Table~\ref{tab:interventions}). Self-consistency and ensembles retain negative mean biases of approximately $-1.3$ to $-1.5$ rating points, while fine-tuned score placement depends on protocol (\S\ref{sec:severity}). The TF--IDF control has small mean bias ($-0.05$) and competitive ordering, yet estimates acceptance at 43.5\% versus the workers' 61.1\%. Matching the mean therefore does not recover the acceptance estimand.

\begin{table}[t]
\centering
\small
\setlength{\tabcolsep}{4pt}
\caption{Medians by configuration group for the 32 near-complete arms ($\geq$99.8\% test coverage); acceptance uses scores $\geq7$. Groups differ in models and protocols, so these are descriptive summaries, not controlled intervention comparisons.}
\label{tab:interventions}
\begin{tabular}{lrrrrrr}
\toprule
Configuration group & $n$ & Pair acc. & Bias & $|$Bias$|$ & Acc.\ (score ${\geq}7$) & Cov. \\
\midrule
proprietary pointwise & 3 & 0.595 & $-2.50$ & 2.50 & 0.117 & 0.66 \\
few-shot/listwise prompting & 8 & 0.629 & $-1.61$ & 1.61 & 0.348 & 0.90 \\
self-consistency & 2 & 0.637 & $-1.53$ & 1.53 & 0.336 & 0.93 \\
ensembling & 4 & 0.643 & $-1.31$ & 1.31 & 0.353 & 0.92 \\
scaffold prompting & 3 & 0.620 & $-0.82$ & 0.82 & 0.497 & 0.89 \\
reasoning/decoding & 3 & 0.643 & $-0.74$ & 0.74 & 0.495 & 0.89 \\
fine-tuned (listwise) & 4 & 0.654 & $-0.62$ & 0.62 & 0.513 & 0.91 \\
open-weight pointwise & 3 & 0.545 & $+0.12$ & 0.91 & 0.784 & 0.53 \\
fine-tuned (pointwise) & 2 & 0.592 & $+1.22$ & 1.22 & 0.903 & 0.49 \\
control: TF-IDF (response only) & 1 & 0.643 & $-0.05$ & 0.05 & 0.435 & 1.00 \\
\midrule
Workers (target raters) & --- & --- & $0$ & 0 & 0.611 & --- \\
\bottomrule
\end{tabular}

\end{table}

\subsection{Scale disagreement propagates to the reported occupational quantity}
\label{sec:consequences}

Workers rate 61.1\% of test responses acceptable at the survey threshold of 7. On the same responses, judge acceptance shares range from 3.0\% for Opus to 97.9\% for near-constant Llama, with a median of 43.9\%; the pointwise fine-tunes yield 89.8\% and 90.8\%. Thus, the reported acceptance share depends strongly on the scoring instrument even with responses held fixed. This comparison measures disagreement with the sampled worker reference (\S\ref{sec:limitations}).

The score differences also change occupation classifications. Workers rate a strict majority of sampled responses acceptable in 68.7\% of the 307 evaluable occupations. Opus does so in 0.3\% $[0.0, 1.0]$ and disagrees with the sampled worker majority in 68.4\% $[63.5, 73.6]$ of occupations. The pointwise fine-tune and near-zero-mean-bias GPT-OSS disagree in 31.3\% and 30.9\%, respectively, compared with 31.3\% for a constant classifier that labels every occupation acceptable (Appendix~\ref{app:occ}). These are disagreements with sampled worker majorities, not errors against an established population classification.

\begin{figure}[t]
\centering
\includegraphics[width=\textwidth]{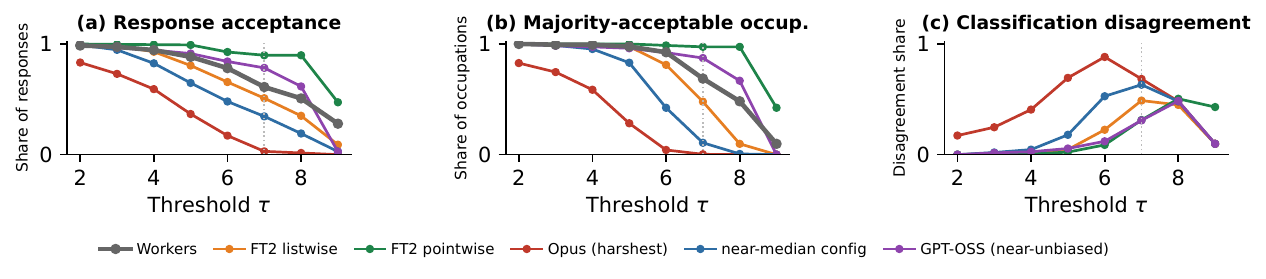}
\caption{Downstream consequences by threshold $\tau$. (a)~Share of responses judged acceptable. (b)~Share of occupations with a strict majority acceptable. (c)~Share of occupations whose classification disagrees with the sampled worker majority.}
\label{fig:consequences}
\end{figure}

Threshold, cell-size, and aggregation-weight sensitivity checks preserve substantial disagreement (Figure~\ref{fig:consequences}; Appendix~\ref{app:occ}). Restricting to the 102 occupations whose worker acceptance share is bootstrap-separated from one half yields disagreement of 15\%--84\% across configurations, with a median of 66\%. This restriction reduces ambiguity around the sampled majority but also changes the subset's composition and does not resolve sparse task coverage or uncertainty about the underlying occupational population.

\subsection{Calibration and label efficiency}
\label{sec:calibration}

Two label-based repairs could in principle restore the scale: recalibrating judge scores against worker labels, and using judge predictions to stretch a small labeled sample. We evaluate both. Task-grouped five-fold isotonic calibration reduces absolute mean bias to at most 0.003 for every near-complete configuration. Predictive agreement with individual worker ratings remains limited: the best cross-validated $R^2$ is 0.085, with residual SD 1.85 compared with worker-rating SD 1.94 (Appendix~\ref{app:calibration}). This result concerns prediction of observed individual ratings; it is not an estimate of agreement with noise-free response quality. Calibration must also match the intended estimand. A score calibrated to predict a mean rating does not directly estimate the probability that the rating is at least 7: conditional-mean predictors shrink dispersion, so hard thresholding under-counts the upper tail of a distribution concentrated near the cutoff. Accordingly, thresholding mean-calibrated scores leaves a median acceptance-share error of 10.7 percentage points, whereas directly calibrating acceptance probabilities reduces pooled error to at most 0.1 percentage points. A fold-fitted constant achieves similarly small pooled error without judge input, so aggregate bias correction alone does not establish useful predictive information (Table~\ref{tab:caltargets}).

We also evaluate whether judge predictions improve estimation of the worker mean or acceptance share when only a task-clustered fraction of test ratings is labeled (prediction-assisted estimation; \citealp{angelopoulos2023ppi}). The cross-fitted construction combines a finite-population, task-clustered residual variance with a heuristic calibrator-fitting term (procedure and coverage assessment: Appendix~\ref{app:calibration}). Across the evaluated conditions, absolute bias is at most 0.013 and empirical coverage is 0.92--0.98. However, estimator SD is not consistently lower than for human-only estimation (median SD ratio 1.00), and the constructed intervals are wider at every studied budget. A tuned-weight heuristic and a TF--IDF predictor provide no consistent improvement in this cross-fitted design. Raw train-frozen scores with a tuned weight yield at most small gains, with SD ratios of 0.96--1.02; these sit at the theoretical floor, since with $R^2\leq0.085$ a tuned linear correction can reduce estimator SD by at most $1-\sqrt{1-R^2}\approx4.3\%$. These results indicate limited label-efficiency gains in this setting; they do not establish that prediction assistance cannot help elsewhere.

\subsection{A replicated ordering--aggregation reversal in one fine-tuned lineage}
\label{sec:reversal}

The clearest ordering--aggregation reversal occurs within the two-epoch fine-tuned lineage. Holding checkpoint and temperature fixed, changing from pointwise to the bundled few-shot/listwise protocol raises response Spearman correlation from 0.214 to 0.263 (difference $+0.049$ $[0.021, 0.078]$) and tie-aware pair accuracy from 0.596 to 0.658. In the same comparison, correlation with worker means falls from 0.215 to 0.096 at the task level ($-0.119$ $[-0.218,-0.022]$) and from 0.254 to 0.087 at the occupation level ($-0.167$ $[-0.296,-0.043]$). All three contrasts satisfy the prespecified validation criteria; validation differences are $+0.033$, $-0.112$, and $-0.108$, respectively, with every interval excluding zero (Table~\ref{tab:reversal}; Appendix~\ref{app:replication}). Aggregate correlations throughout should be read against the low reliability of the worker cell means (Appendix~\ref{app:reliability}).

The broader matched comparisons delimit this result. In the base-model $2{\times}2$, the estimated presentation/schema effect on response correlation exceeds the exemplar effect on both test ($+0.082$ versus $+0.028$) and validation ($+0.092$ versus $+0.019$), while aggregate effects are small and inconsistent across splits (Appendix~\ref{app:pairs}). The strong replicated reversal therefore supports a finding about the evaluated fine-tuned lineage and protocol bundle, not a general claim that listwise presentation reduces aggregate validity.

\looseness=-1 The distribution of score variance is consistent with more rank-oriented listwise scoring: the share lying across tasks falls from 29.5\% to 11.8\% and across occupations from 20.7\% to 7.8\%, while the share aligned with response-generator identity rises from 13.2\% to 34.6\% (position controls remove only 14\% of that shift; Appendix~\ref{app:mechanism}). Listwise output approaches a within-set rank code: only 18.1\% of its variance lies between task instances, and the judge's own within-instance rank explains 90.1\% of the rest; scores are monotone in their own ranks, so the evidence is the pointwise contrast (37.8\% and 69.4\%). A rank code carries little between-cell location, consistent with the protocol-dependent listwise ``bias'' and the decline in task- and occupation-level agreement. Permuting each instance's pointwise values into the listwise order recovers part of the local gain with every cell mean unchanged, so ordering and location are separable (Appendix~\ref{app:hybrid}).

\section{Discussion}

The audit's practical implication concerns validation practice, not any single judge configuration. Judge validation should match the reported quantity: acceptance probabilities or shares for acceptance estimates, and task or occupation means for aggregates. Ranking agreement alone does not establish these properties. The distinction follows from rank invariance (\S\ref{sec:metrics}), but the observed reversal concerns one fine-tuned protocol bundle; the distributional diagnostics do not identify a general causal mechanism.

Calibration should be evaluated on held-out labels with simple label-fitted controls. Prediction assistance should be compared with human-only estimation for precision and interval coverage at the same label budget. The small gains here show that ranking performance and mean-bias correction do not establish label efficiency.

\paragraph{Scope and limitations.}
\label{sec:limitations}
The corpus covers English-language text outputs for U.S. occupational tasks screened for text-based suitability, rated by workers with self-reported relevant experience; the sample is not employment-weighted, and acceptance prevalence depends on the generator mix. The worker ratings define the audit's target, not an objective standard: each response has one rating, so we cannot determine whether a disagreement reflects the worker, the judge, or both, and the low split-half estimates do not quantify reliability at every level (Appendix~\ref{app:reliability}). Test-split analyses remain exploratory; prespecified validation covers the named reversal, not every audit result. The strong aggregate reversal occurs in one fine-tuned lineage under a bundled protocol change (\S\ref{sec:reversal}); response order was not randomized, limiting position diagnostics (Appendix~\ref{app:mechanism}); and generalization to other work settings, modalities, or population capability estimates requires additional evidence.

\section{Conclusion}

\looseness=-1 Agreement with workers on response ordering did not ensure agreement on score levels, acceptance rates, or occupational aggregates. A protocol change in one fine-tuned lineage improved ordering while reducing aggregate agreement. Calibration corrected the quantity it targeted, while prediction-assisted estimation yielded at most small precision gains. Occupational evaluations should therefore validate judges against the quantities they intend to report (Appendix~\ref{app:standard}). Extending these findings to other occupational settings will require independent corpora and repeated human ratings.

\bibliography{references}
\bibliographystyle{iclr2027_conference}

\appendix
\raggedbottom

\section{Reporting standard}
\label{app:standard}

For evaluations that aggregate LLM-judge scores:
\begin{enumerate}
  \item \textbf{Decision level.} Name the target (response, item, domain/occupation, population, or system) before selecting the judge.
  \item \textbf{Ordering with ties and coverage.} Report tie-aware accuracy, accuracy conditional on a decision, and the share of evaluable comparisons receiving a decision. For listwise or comparative protocols, randomize or counterbalance presentation order and report position sensitivity.
  \item \textbf{Score level and distribution.} Report signed bias with clustered uncertainty, the score-SD ratio, the full distribution, and threshold-crossing rates across a threshold sweep.
  \item \textbf{Simple controls.} Include train-fitted lexical predictors and group-mean controls to expose stable priors that can drive apparent agreement.
  \item \textbf{Level-matched uncertainty and target reliability.} Cluster at the assignment unit, propagate within-cell sampling error for hierarchical means, and report the reliability of the human quantity being predicted.
  \item \textbf{Separated configuration and confirmation data.} When configurations are tuned or repeatedly inspected, re-estimate the selected comparison on excluded data under prespecified directions or thresholds.
\end{enumerate}

\section{Configuration registry with severity and ordering metrics}
\label{app:registry}

\begin{figure}[H]
\centering
\includegraphics[width=\textwidth]{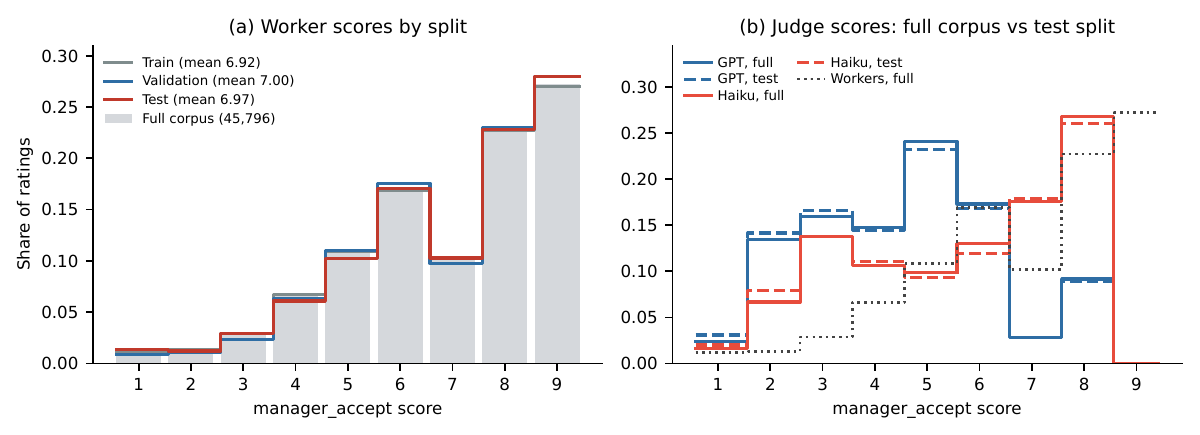}
\caption{Split representativeness. (a)~Worker \texttt{manager\_accept} shares by split against the full corpus: split means are 6.92/7.00/6.97, and the largest per-score share difference between the test split and the full corpus is 0.008. (b)~The two configurations run on the full corpus reproduce their test-split distributions (GPT-5.4-mini mean 4.54 versus 4.47; Haiku 5.64 versus 5.56), and both assign a score of 9 to fewer than 0.05\% of the 45{,}796 rows.}
\label{fig:appsplits}
\end{figure}

Table~\ref{tab:registry} lists all 33 judge configurations and the six train-fitted controls with row coverage (Cover.), mean bias (95\% task-clustered CI), acceptance share at scores $\geq7$ (Acc.; workers: 0.611), tie-aware pair accuracy (PairAcc), and decisive-pair coverage (PairCov); Table~\ref{tab:namekey} decodes the configuration names; and Table~\ref{tab:splits} details the splits. The reasoning variant with 22.3\% row coverage is reported on covered rows and excluded from cross-configuration statistics. Figure~\ref{fig:appsplits} checks split representativeness against the full corpus.

\begin{table}[H]
\centering
\scriptsize
\setlength{\tabcolsep}{3pt}
\caption{Full registry. ``Present.''\ is the presentation protocol; naming tokens are decoded in Table~\ref{tab:namekey}, and column quantities are defined in \S\ref{sec:metrics}. Bias intervals use 1{,}000 task-clustered bootstrap draws.}
\label{tab:registry}
\begin{tabular}{llllllll}
\toprule
Configuration & Family & Present. & Cover. & Bias [95\% CI] & Acc.\ (score ${\geq}7$) & PairAcc & PairCov \\
\midrule
Claude Opus 4.7 pointwise & Claude & pointwise & 1.000 & $-3.23$ {\scriptsize$[-3.36,-3.11]$} & 0.030 & 0.595 & 0.618 \\
Haiku pointwise & Claude & pointwise & 1.000 & $-1.41$ {\scriptsize$[-1.56,-1.28]$} & 0.439 & 0.600 & 0.709 \\
Haiku fs+lw & Claude & listwise & 1.000 & $-1.33$ {\scriptsize$[-1.43,-1.22]$} & 0.377 & 0.632 & 0.929 \\
\midrule
GPT pointwise & GPT & pointwise & 1.000 & $-2.50$ {\scriptsize$[-2.63,-2.37]$} & 0.117 & 0.568 & 0.657 \\
GPT fs+lw & GPT & listwise & 1.000 & $-2.47$ {\scriptsize$[-2.60,-2.35]$} & 0.214 & 0.601 & 0.869 \\
\midrule
GPT-OSS fs+lw & GPT-OSS & listwise & 1.000 & $-1.49$ {\scriptsize$[-1.61,-1.37]$} & 0.337 & 0.632 & 0.855 \\
GPT-OSS pointwise & GPT-OSS & pointwise & 1.000 & $+0.12$ {\scriptsize$[+0.01,+0.22]$} & 0.784 & 0.545 & 0.531 \\
\midrule
DeepSeek-V4-Pro fs+lw & DeepSeek & listwise & 1.000 & $-2.09$ {\scriptsize$[-2.21,-1.99]$} & 0.283 & 0.619 & 0.923 \\
\midrule
Llama-3.3-70B pointwise & Llama & pointwise & 1.000 & $+0.91$ {\scriptsize$[+0.83,+0.99]$} & 0.979 & 0.507 & 0.051 \\
\midrule
Qwen few-shot & Qwen-base & pointwise & 1.000 & $-2.01$ {\scriptsize$[-2.16,-1.87]$} & 0.315 & 0.594 & 0.798 \\
Qwen fs(10ex)+lw & Qwen-base & listwise & 1.000 & $-1.61$ {\scriptsize$[-1.73,-1.50]$} & 0.361 & 0.631 & 0.922 \\
Qwen listwise & Qwen-base & listwise & 1.000 & $-1.61$ {\scriptsize$[-1.74,-1.48]$} & 0.381 & 0.627 & 0.856 \\
Qwen fs+lw & Qwen-base & listwise & 1.000 & $-1.58$ {\scriptsize$[-1.69,-1.47]$} & 0.360 & 0.634 & 0.921 \\
SelfConsist average (Qwen 5x) & Qwen-base & listwise & 1.000 & $-1.54$ {\scriptsize$[-1.65,-1.43]$} & 0.307 & 0.639 & 0.972 \\
SelfConsist majority (Qwen 5x) & Qwen-base & listwise & 1.000 & $-1.53$ {\scriptsize$[-1.63,-1.42]$} & 0.366 & 0.635 & 0.896 \\
Qwen pointwise & Qwen-base & pointwise & 1.000 & $-1.09$ {\scriptsize$[-1.25,-0.94]$} & 0.476 & 0.558 & 0.754 \\
Qwen analogical+lw & Qwen-base & listwise & 1.000 & $-0.98$ {\scriptsize$[-1.07,-0.87]$} & 0.449 & 0.628 & 0.892 \\
Qwen ranked+lw & Qwen-base & listwise & 0.999 & $-0.82$ {\scriptsize$[-0.90,-0.73]$} & 0.497 & 0.620 & 0.983 \\
Qwen rubric+fs+lw & Qwen-base & listwise & 1.000 & $-0.72$ {\scriptsize$[-0.84,-0.61]$} & 0.529 & 0.615 & 0.835 \\
\midrule
Qwen-FT-2ep 5ex think/t1 & Qwen-FT & listwise & 0.223 & $-0.76$ {\scriptsize$[-0.99,-0.54]$} & 0.485 & 0.659 & 0.940 \\
Qwen-FT-2ep 5ex noThink/t0.7 & Qwen-FT & listwise & 1.000 & $-0.75$ {\scriptsize$[-0.84,-0.64]$} & 0.495 & 0.643 & 0.888 \\
Qwen-FT-2ep 10ex noThink/t0.7 & Qwen-FT & listwise & 1.000 & $-0.74$ {\scriptsize$[-0.83,-0.64]$} & 0.494 & 0.645 & 0.896 \\
Qwen-FT-2ep 10ex & Qwen-FT & listwise & 1.000 & $-0.67$ {\scriptsize$[-0.75,-0.57]$} & 0.507 & 0.654 & 0.923 \\
Qwen-FT-2ep fs+lw & Qwen-FT & listwise & 1.000 & $-0.65$ {\scriptsize$[-0.73,-0.55]$} & 0.510 & 0.658 & 0.912 \\
Qwen-FT-2ep analog noThink/t0.7 & Qwen-FT & listwise & 1.000 & $-0.60$ {\scriptsize$[-0.69,-0.51]$} & 0.523 & 0.628 & 0.819 \\
Qwen-FT-1ep fs+lw & Qwen-FT & listwise & 1.000 & $-0.59$ {\scriptsize$[-0.68,-0.50]$} & 0.517 & 0.655 & 0.914 \\
Qwen-FT-2ep analogical & Qwen-FT & listwise & 1.000 & $-0.52$ {\scriptsize$[-0.61,-0.43]$} & 0.554 & 0.639 & 0.757 \\
Qwen-FT-2ep pointwise & Qwen-FT & pointwise & 1.000 & $+1.21$ {\scriptsize$[+1.13,+1.29]$} & 0.898 & 0.596 & 0.510 \\
Qwen-FT-1ep pointwise & Qwen-FT & pointwise & 1.000 & $+1.24$ {\scriptsize$[+1.16,+1.32]$} & 0.908 & 0.588 & 0.479 \\
\midrule
Ensemble3 average (fs+lw) & Ensemble & ensemble & 1.000 & $-1.47$ {\scriptsize$[-1.58,-1.36]$} & 0.304 & 0.644 & 0.953 \\
Ensemble3 majority (fs+lw) & Ensemble & ensemble & 1.000 & $-1.43$ {\scriptsize$[-1.55,-1.33]$} & 0.361 & 0.638 & 0.877 \\
EnsHQF average (H+Q+FT) & Ensemble & ensemble & 1.000 & $-1.19$ {\scriptsize$[-1.28,-1.09]$} & 0.346 & 0.653 & 0.961 \\
EnsHQF majority (H+Q+FT) & Ensemble & ensemble & 1.000 & $-1.18$ {\scriptsize$[-1.28,-1.09]$} & 0.407 & 0.643 & 0.897 \\
\midrule
Control: train mean & --- & constructed & 1.000 & $-0.05$ {\scriptsize$[-0.13,+0.04]$} & 0.000 & 0.500 & 0.000 \\
Control: response position & --- & constructed & 1.000 & $-0.05$ {\scriptsize$[-0.13,+0.04]$} & 0.400 & 0.531 & 1.000 \\
Control: occupation mean & --- & constructed & 1.000 & $-0.04$ {\scriptsize$[-0.12,+0.05]$} & 0.435 & 0.500 & 0.000 \\
Control: length + position & --- & constructed & 1.000 & $-0.05$ {\scriptsize$[-0.13,+0.03]$} & 0.407 & 0.592 & 1.000 \\
Control: TF--IDF response only & --- & constructed & 1.000 & $-0.05$ {\scriptsize$[-0.13,+0.02]$} & 0.435 & 0.643 & 1.000 \\
Control: TF--IDF full context & --- & constructed & 1.000 & $-0.05$ {\scriptsize$[-0.12,+0.03]$} & 0.438 & 0.634 & 1.000 \\
\bottomrule
\end{tabular}

\end{table}

\begin{table}[H]
\centering
\footnotesize
\setlength{\tabcolsep}{5pt}
\caption{Configuration naming key for Table~\ref{tab:registry}, Figure~\ref{fig:severity}a, and Figure~\ref{fig:appdist}.}
\label{tab:namekey}
\begin{tabular}{@{}>{\raggedright\arraybackslash}p{2.6cm}>{\raggedright\arraybackslash}p{10.3cm}@{}}
\toprule
Token & Meaning \\
\midrule
pointwise / lw & one response per prompt / all responses of a task instance (typically five) scored together in one prompt (\S\ref{sec:data}) \\
fs & few-shot exemplars in the prompt; \emph{5ex}/\emph{10ex} give the count where it is varied \\
rubric, ranked, analogical & alternative prompt scaffolds \\
think / noThink & hybrid reasoning enabled / disabled \\
t0.7, t1 & decoding temperature (all other arms decode at 0) \\
5x & five sampled judgments per response (self-consistency), combined by average or majority \\
Ensemble3, EnsHQF & three-judge ensembles of listwise members; H+Q+FT = Haiku, Qwen, and the fine-tuned checkpoint \\
FT-1ep / FT-2ep & LoRA checkpoint after one / two training epochs (Appendix~\ref{app:ft}) \\
Control: & train-fitted non-LLM predictor (\S\ref{sec:data}) \\
\bottomrule
\end{tabular}
\end{table}

\begin{table}[H]
\centering
\small
\setlength{\tabcolsep}{5pt}
\caption{Dataset splits. Task IDs and workers are disjoint across splits; occupations and response-generator variants overlap.}
\label{tab:splits}
\begin{tabular}{lrrrrrr}
\toprule
Split & Ratings & Task IDs & Workers & Occup. & Families & Generators \\
\midrule
Train & 36,649 & 4,119 & 7,339 & 680 & 22 & 47 \\
Validation & 4,646 & 515 & 931 & 311 & 22 & 47 \\
Test & 4,501 & 515 & 901 & 307 & 22 & 47 \\
\bottomrule
\end{tabular}
\end{table}

\section{Prompt templates (abridged)}
\label{app:prompts}

The two templates below anchor the primary contrast of \S\ref{sec:reversal}; both are abridged, with runner-filled placeholders in braces and elisions marked by [...]. The remaining six variants (zero-shot listwise, rubric, ranked, analogical, ten-exemplar few-shot/listwise, and few-shot pointwise) modify these skeletons.

\paragraph{Pointwise (original 35-field survey schema).}
{\footnotesize
\begin{verbatim}
SYSTEM  You are a professional worker completing a workplace survey. You will
be told your occupation and years of experience [...] Your response must be a
single JSON object that follows the exact schema provided in the user message.
Output ONLY the JSON object. Use exactly the field names shown. [...]

USER  You are a {selected_occupation} with {work_experience} of work experience.
Your manager has assigned you the following task. One of the general
responsibilities of your job is: "{task_desc}"
Here is the specific task you have been assigned: {task_instance_text}
A colleague has produced a response to this task. Here is response
#{response_index}: {response_full_text}
Now answer the following survey questions. Provide your answers as a single
JSON object using exactly this schema:
{ "task_understandability": "", ...        [33 further survey fields elided]
  "resp_manager_accept": 0,
  // Overall, how would a manager view this response?
  // 1 = Not useful: Requires complete rework (started over from scratch)
  // 2 = Not useful: Requires extensive rework
  // 3 = Not useful: Requires substantial rework
  // 4 = Useful with edits: Requires major edits
  // 5 = Useful with edits: Requires moderate edits
  // 6 = Useful with edits: Requires minor edits
  // 7 = Useful as is: Requires no edits to be minimally sufficient
  // 8 = Useful as is: Requires no edits to be of average quality
  // 9 = Useful as is: Requires no edits to be of superior quality
  ... }
\end{verbatim}
}

\paragraph{Few-shot listwise (manager rating only).}
{\footnotesize
\begin{verbatim}
SYSTEM  You are a professional worker completing a workplace survey. [...] To
help you judge consistently, you will first be given the rating scale, then a
worked example from your field: a task with several real responses and the
manager rating each one actually received. Use the example to calibrate how
you apply the scale. You will then be shown several candidate responses to a
NEW task; read all of them before judging, and rate each relative to the
others. [...]

USER  You are a {selected_occupation} with {work_experience} of work experience.
You will judge how a manager would view a colleague's response to a task,
using the following 1-9 scale:            [the nine anchors as shown above]
WORKED EXAMPLE (to calibrate the scale) [...] "{example_task_desc}"
{example_task_instance_text}
Example responses and the manager rating each received: {anchor_block}
NOW RATE A NEW TASK
Apply the same 1-9 scale, calibrated to the example above. Below are several
candidate responses to a new task assigned to you. Read ALL of them before
answering, then judge them together. [...] "{task_desc}" {task_instance_text}
Candidate responses to judge: {responses_block}
Provide your answer as a single JSON object using exactly this schema:
{ "ratings": [ { "response_index": 0, "resp_manager_accept": 0 } ] }
\end{verbatim}
}

The few-shot anchor is a worked example from the same occupation in the training split (fallback: the same job family), shown with all of its responses and the manager rating each received, in shuffled order.

\section{Score distributions and score placement}
\label{app:distributions}

\begin{figure}[H]
\centering
\includegraphics[width=\textwidth]{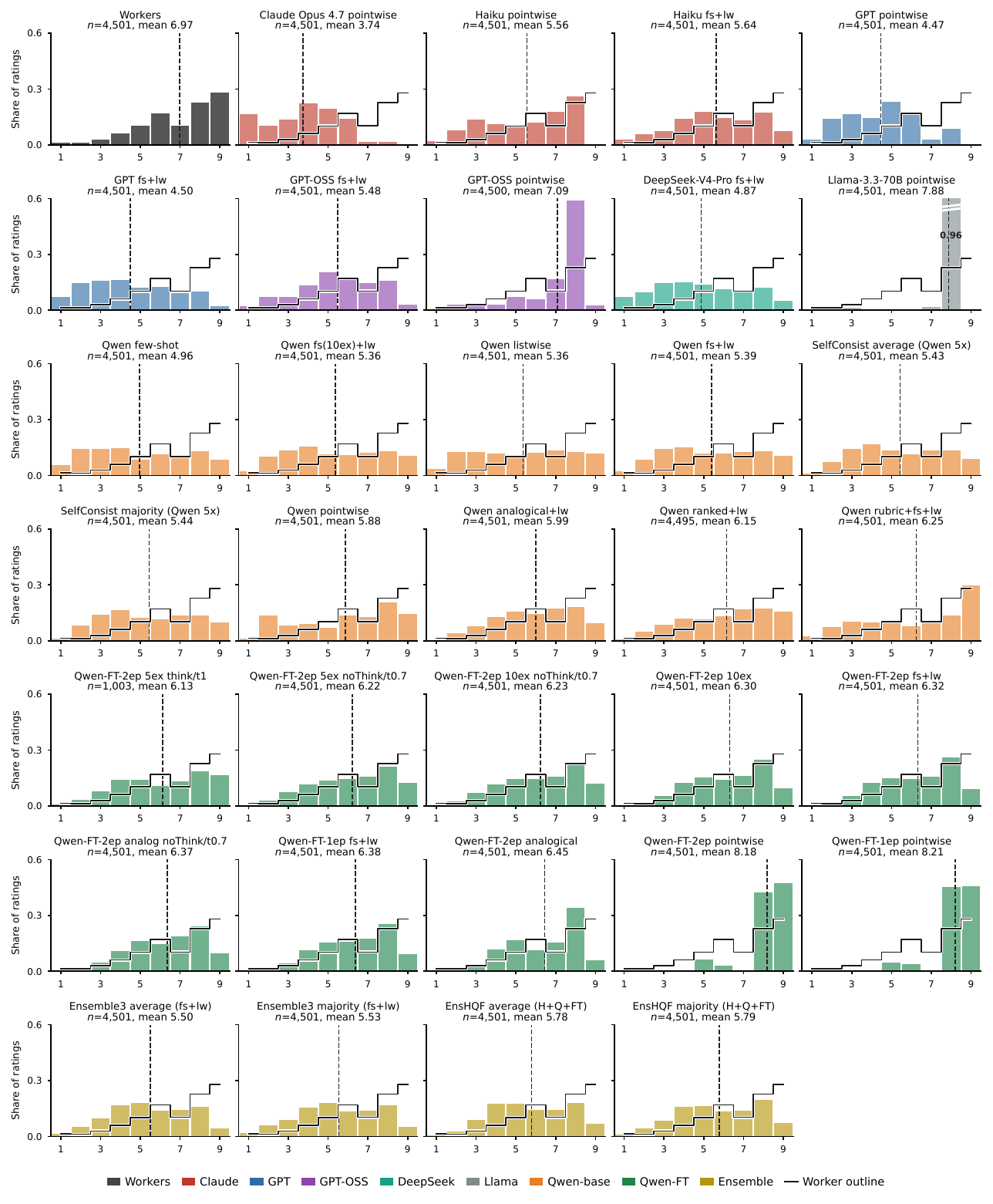}
\caption{Score distributions for the workers (top left) and all 33 configurations on the test split, grouped and colored by model family (Figure~\ref{fig:severity}b palette; legend below). Bars are shares of ratings on a common axis capped at 0.6; the one clipped bar is break-marked and annotated with its value, and the cased dark line in each judge panel is the workers' distribution. Panel headers give valid rows and the mean (workers: 6.97); the dashed line marks each panel's mean. No configuration reproduces the workers' right-shifted distribution: most judges shift mass left of the worker outline, the pointwise fine-tunes overshoot onto 8--9, and Llama collapses to a near-constant 8.}
\label{fig:appdist}
\end{figure}

\begin{figure}[H]
\centering
\includegraphics[width=\textwidth]{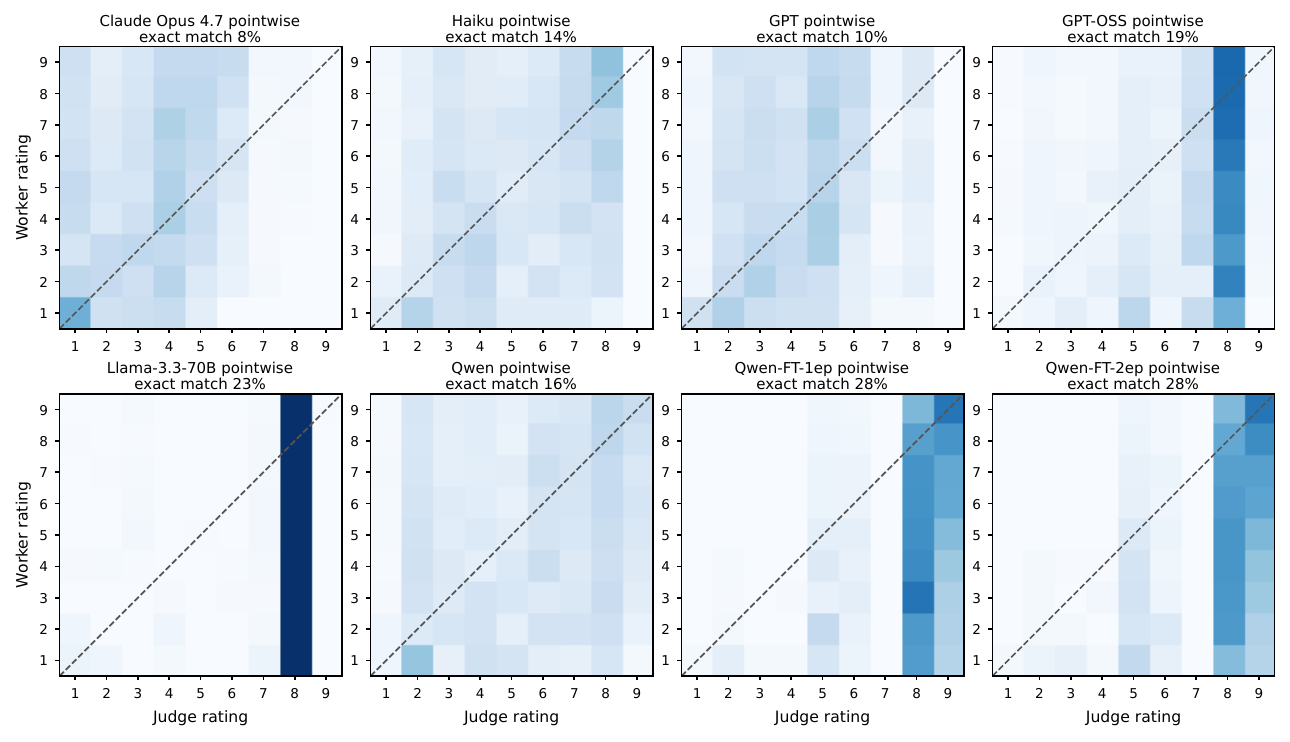}
\caption{Conditional distributions of judge scores (columns) given worker scores (rows), normalized within each worker-score row, for eight pointwise configurations on the test split. Panel titles report exact-match rates. The distributions show differences in score placement and concentration; Llama assigns nearly constant scores. Agreement in within-instance response ordering is evaluated separately using the metrics in \S\ref{sec:ordering}.}
\label{fig:appconf}
\end{figure}

\begin{figure}[H]
\centering
\includegraphics[width=0.87\textwidth]{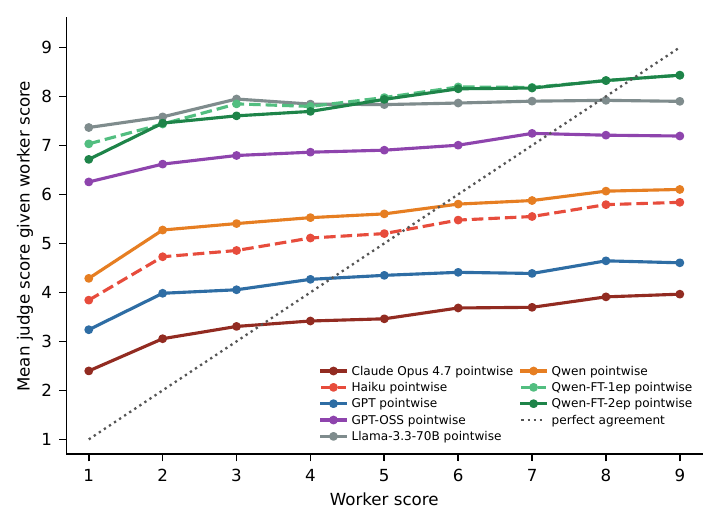}
\caption{Mean judge score conditional on each worker score for eight pointwise configurations on the test split. The diagonal marks equality between the conditional mean judge score and the worker score. Conditional means vary less than the full worker-score range, and their levels differ substantially across configurations. This descriptive view complements the marginal distributions in Figure~\ref{fig:appdist} and the ordering metrics in \S\ref{sec:ordering}.}
\label{fig:appbucket}
\end{figure}

Figure~\ref{fig:appdist} compares the \texttt{manager\_accept} score distribution of each configuration with the worker distribution, supplementing Figure~\ref{fig:scale}b and \S\ref{sec:severity}. Figure~\ref{fig:appconf} shows row-normalized joint score distributions for eight pointwise configurations, making systematic shifts and concentration visible. Figure~\ref{fig:appbucket} summarizes the conditional mean judge score at each worker rating: these means vary over a narrower range than the worker scale and do not track the diagonal across its full range. These plots describe score placement; within-instance ordering agreement is quantified separately in \S\ref{sec:ordering}.

\section{Matched protocol pairs and the \texorpdfstring{$2\times2$}{2x2} decomposition}
\label{app:pairs}

\begin{table}[H]
\centering
\footnotesize
\setlength{\tabcolsep}{2pt}
\caption{Matched pairs with 95\% task-clustered intervals; boldface marks estimates whose intervals exclude zero. Local ordering improves in all six pairs. Statistically clear aggregate declines are confined to the two checkpoints of one LoRA lineage, which differ by a single training epoch and are not independent. The FT2 row reports the canonical preregistered intervals of Table~\ref{tab:reversal}; the remaining rows use an earlier matched-pair bootstrap draw set.}
\label{tab:multiverse}
\begin{tabular}{lrrr}
\toprule
Pointwise $\to$ few-shot/listwise & $\Delta$ response $\rho$ & $\Delta$ task $\rho$ & $\Delta$ occ.\ $\rho$ \\
\midrule
Qwen base & \textbf{$+0.110$} [0.077,0.144] & $-0.006$ [$-0.079$,0.072] & $-0.025$ [$-0.079$,0.062] \\
Haiku & \textbf{$+0.070$} [0.042,0.096] & $-0.011$ [$-0.065$,0.044] & $-0.007$ [$-0.058$,0.047] \\
GPT-OSS & \textbf{$+0.109$} [0.078,0.142] & $-0.003$ [$-0.078$,0.072] & $+0.029$ [$-0.055$,0.090] \\
GPT & \textbf{$+0.060$} [0.036,0.085] & $+0.031$ [$-0.017$,0.077] & $+0.013$ [$-0.026$,0.072] \\
Qwen FT1 & \textbf{$+0.043$} [0.014,0.072] & \textbf{$-0.116$} [$-0.207$,$-0.023$] & \textbf{$-0.145$} [$-0.224$,$-0.041$] \\
Qwen FT2 & \textbf{$+0.049$} [0.021,0.078] & \textbf{$-0.119$} [$-0.218$,$-0.022$] & \textbf{$-0.167$} [$-0.296$,$-0.043$] \\
\bottomrule
\end{tabular}
\end{table}

\begin{table}[H]
\centering
\footnotesize
\setlength{\tabcolsep}{5pt}
\caption{\texorpdfstring{$2\times2$}{2x2} on base Qwen3.5-9B, test and validation. Mean response-level effects, computed from unrounded cell means: presentation/schema $+0.082$ (test) and $+0.092$ $[0.068, 0.115]$ (validation); exemplars $+0.028$ and $+0.019$ $[0.001, 0.038]$. Occupation-level factor effects are small and change sign across splits.}
\label{tab:factorial}
\begin{tabular}{llrrr@{\qquad}rrr}
\toprule
& & \multicolumn{3}{c}{Test} & \multicolumn{3}{c}{Validation} \\
\cmidrule(lr){3-5}\cmidrule(lr){6-8}
Presentation & Exemplars & Resp.\ $\rho$ & Task $\rho$ & Occ.\ $\rho$ & Resp.\ $\rho$ & Task $\rho$ & Occ.\ $\rho$ \\
\midrule
Pointwise & zero-shot & 0.109 & 0.067 & 0.075 & 0.088 & 0.051 & 0.007 \\
Pointwise & few-shot  & 0.138 & 0.087 & 0.058 & 0.124 & 0.018 & $-0.066$ \\
Listwise  & zero-shot & 0.191 & 0.087 & 0.036 & 0.196 & 0.092 & 0.015 \\
Listwise  & few-shot  & 0.219 & 0.060 & 0.049 & 0.200 & 0.056 & $-0.049$ \\
\bottomrule
\end{tabular}
\end{table}

\section{Validation-split replications}
\label{app:replication}

\paragraph{Protocol reversal.} Table~\ref{tab:reversal} reports the preregistered replication: 4{,}646 ratings, disjoint task IDs, task instances, and workers; directions and thresholds recorded before the runs; both arms at complete coverage. The preregistered criterion was the validation point estimate falling beyond the stated threshold; the reported intervals additionally exclude zero, though they do not establish effects as large as the thresholds. Levels on validation: response 0.216 $\to$ 0.250; task 0.241 $\to$ 0.129; occupation 0.201 $\to$ 0.093. Differences here and in Tables~\ref{tab:factorial} and~\ref{tab:reversal} are computed from unrounded values.

\begin{table}[H]
\centering
\small
\setlength{\tabcolsep}{5pt}
\caption{Same-checkpoint protocol reversal (few-shot/listwise minus pointwise, Qwen-FT-2ep, temperature~0), task-clustered intervals. Thresholds were fixed before the validation runs; all three replicate.}
\label{tab:reversal}
\begin{tabular}{lrrr}
\toprule
Level & Test $\Delta$ [95\% CI] & Validation $\Delta$ [95\% CI] & Pre-registered threshold \\
\midrule
Response $\rho$ & $+0.049$ $[+0.021, +0.078]$ & $+0.033$ $[+0.003, +0.062]$ & $\geq +0.025$ \\
Task $\rho$ & $-0.119$ $[-0.218, -0.022]$ & $-0.112$ $[-0.185, -0.032]$ & $\leq -0.05$ \\
Occupation $\rho$ & $-0.167$ $[-0.296, -0.043]$ & $-0.108$ $[-0.196, -0.006]$ & $\leq -0.10$ \\
\bottomrule
\end{tabular}
\end{table}

\paragraph{Severity (validation robustness check, not preregistered).} Mean bias on validation for the arms with validation predictions, against their test values: Haiku pointwise $-1.41$ (test $-1.41$); GPT-5.4-mini pointwise $-2.49$ ($-2.50$); Qwen base pointwise $-1.07$ ($-1.09$); Qwen few-shot pointwise $-1.99$ ($-2.01$); Qwen listwise zero-shot $-1.71$ ($-1.61$); Qwen few-shot/listwise $-1.61$ ($-1.58$); FT2 pointwise $+1.26$ ($+1.21$); FT2 few-shot/listwise $-0.70$ ($-0.65$). Validation acceptance at scores $\geq7$: workers 0.607; FT2 pointwise 0.920; FT2 few-shot/listwise 0.498.

\section{Protocol diagnostics}
\label{app:mechanism}

\begin{table}[H]
\centering
\small
\setlength{\tabcolsep}{5pt}
\caption{One-way variance shares for the primary matched comparison (shares overlap; do not sum). Task-clustered bootstrap intervals.}
\label{tab:variance}
\begin{tabular}{lrrrr}
\toprule
Grouping & Worker $V_G$ & Pointwise $V_G$ & Listwise $V_G$ & $\Delta V_G$ [95\% CI] \\
\midrule
Task & .227 & .295 & .118 & $-$.177 [$-$.212,$-$.143] \\
Occupation & .145 & .207 & .078 & $-$.128 [$-$.181,$-$.111] \\
Job family & .027 & .072 & .016 & $-$.056 [$-$.085,$-$.041] \\
Generator & .058 & .132 & .346 & $+$.214 [.186,.239] \\
\bottomrule
\end{tabular}
\end{table}

\paragraph{Position control.} Generator identity is associated with response index ($\chi^2=1119$, $\mathrm{df}=160$, Cram\'er's $V=0.25$), and listwise judges receive responses in index order. Removing an additive response-index effect from every score column leaves the generator-share increase at $+0.182$ $[0.154, 0.206]$, which is 86\% of the uncontrolled $+0.212$ on the same rows (Table~\ref{tab:variance} reports $+0.214$ on all rows), while the task and occupation contractions are essentially unchanged ($-0.173$ and $-0.126$). Response index carries 5.7\% of listwise score variance versus 0.9\% pointwise and 0.8\% in worker ratings. Because order was not randomized, these are observational controls, not causal position effects.

\paragraph{Scale-aligned diagnostics.} After matching each judge's global mean and SD to the workers', task-cell dispersion relative to workers contracts from 1.11 (pointwise) to 0.71 (listwise) and occupation-cell dispersion from 1.09 to 0.70; slopes of judge on worker group means fall from 0.262 to 0.084 (task) and 0.301 to 0.072 (occupation). Task and occupation offset RMS \emph{decline} ($-0.143$ and $-0.093$) while generator offset RMS rises ($+0.414$ $[0.314, 0.428]$): the failure is compression of occupational structure plus generator-linked variance, not larger group intercepts.

\paragraph{Rank determinism.} For listwise FT2, 18.1\% of score variance lies between task instances and the judge's own within-instance rank explains 90.1\% of within-instance variance; for pointwise FT2 the figures are 37.8\% and 69.4\%. Scores are monotone in their own ranks by construction, so the informative quantity is the pointwise--listwise contrast, consistent with a rank-oriented response pattern.

\paragraph{Length association.} Conditional on worker score, response index, and generator, the listwise FT2 judge assigns $+0.538$ $[0.455, 0.622]$ rating points per SD of log response length, an observational association indicating variation in the judge score not explained by the included controls.

\begin{figure}[H]
\centering
\includegraphics[width=\textwidth]{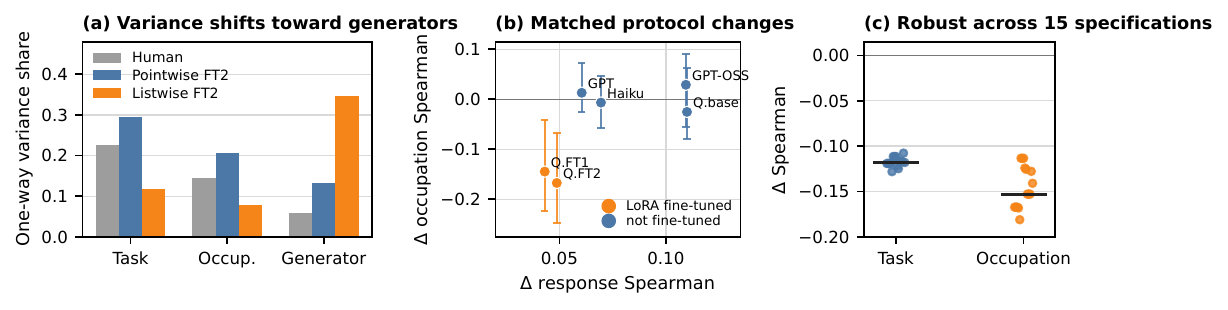}
\caption{Variance patterns and scope of the protocol reversal (\S\ref{sec:reversal}). (a)~In the primary fine-tuned comparison, the few-shot/listwise scores have smaller task- and occupation-associated variance shares and a larger generator-associated share than the pointwise scores; the one-way shares are descriptive and non-additive. (b)~All six matched protocol changes improve response correlation; task-clustered intervals for the occupation-level difference exclude zero only for the two LoRA checkpoints (orange). (c)~The primary task- and occupation-level declines persist across the 15 aggregation specifications; horizontal bars mark medians.}
\label{fig:appmech}
\end{figure}

\paragraph{Aggregation robustness.} Across mean/median/10\%-trimmed aggregation and minimum cell sizes $\{1,3,5,8,10\}$, all 15 task deltas ($-0.128$ to $-0.108$) and all 15 occupation deltas ($-0.181$ to $-0.113$) are negative, as is every leave-one-family-out and leave-one-generator-group-out estimate. Figure~\ref{fig:appmech} summarizes these diagnostics and their scope.

\section{Judge-selection experiment}
\label{app:selection}

In each of 1{,}000 replicates we randomly split the 515 test task IDs in half, compute selection metrics for the 32 near-complete configurations on the selection half, and evaluate the selected configuration on the disjoint evaluation half. Criteria: highest tie-aware pair accuracy; lowest absolute acceptance-share error; lowest absolute mean bias; and highest pair accuracy among configurations within five points of the worker acceptance share (falling back to the acceptance criterion when no configuration qualifies). Because every configuration was previously inspected on the test split, this is a descriptive exercise. Pair accuracy selects the two-epoch listwise fine-tune in 88\% of replicates; the acceptance criterion selects the fine-tuned analogical arm in 99\%; the bias criterion selects GPT-OSS pointwise in every replicate. The combined criterion's eligibility set is empty in 72\% of replicates (no configuration comes within five points of the worker share on the selection half), so it usually reduces to the acceptance fallback and its rows match the acceptance criterion in Table~\ref{tab:selection}.

\begin{table}[H]
\centering
\small
\setlength{\tabcolsep}{5pt}
\caption{Held-out performance by selection criterion over 1{,}000 task half-splits (means; brackets are the 2.5--97.5 percentile range of the acceptance error). Selecting for the reported estimand roughly halves the acceptance error at a cost of 0.018 pair accuracy; selecting for mean bias alone is worst on the acceptance share.}
\label{tab:selection}
\begin{tabular}{lrrr}
\toprule
Selection criterion & $|$Acc.\ err$|$ (pct.\ pts) & $|$Bias$|$ & Pair acc. \\
\midrule
Pair accuracy & 10.5 [7.7, 24.1] & 0.66 & 0.657 \\
Acceptance-share error & 5.8 [3.6, 8.3] & 0.53 & 0.639 \\
Mean-bias magnitude & 17.3 [14.4, 19.9] & 0.12 & 0.546 \\
Pair accuracy within 5 pts of acceptance & 5.8 [3.6, 8.3] & 0.53 & 0.639 \\
\bottomrule
\end{tabular}

\end{table}

\section{Calibration and prediction-powered inference}
\label{app:calibration}

\paragraph{Estimand and procedure.} The target is the finite audited test sample: its mean rating and acceptance share, with ratings equally weighted. Labeled sets are whole task clusters sampled uniformly without replacement, so the human-only standard error carries the finite-population correction $\sqrt{1-k/515}$. The isotonic calibrator is task-level cross-fitted within the labeled set (five folds): each labeled row's rectifier residual comes from a calibrator fit on the other folds, and unlabeled predictions average the fold models. Intervals combine a task-clustered, finite-population rectifier variance with a fold-dispersion term for calibrator fitting; the fold models share training data, so this added term is a heuristic rather than a derived variance component, and the empirical coverage in Table~\ref{tab:ppi} assesses the construction within this simulation only (Monte Carlo standard error about 0.01 per cell). The tuned weight is likewise a plug-in heuristic computed from labeled cluster sums. An in-sample variant that fits and rectifies on the same labeled rows makes the isotonic rectifier zero by construction (isotonic fits preserve the training mean) and overstates efficiency; we therefore report the cross-fitted design and the frozen-predictor contrast below.

\paragraph{Frozen-predictor contrast.} The raw judge scores and the train-fitted TF--IDF control are frozen predictors: no test-split label touches them. The classic estimator $\bar f + \overline{(y-f)}_{L}$ therefore involves no calibrator fitting, and its task-clustered, finite-population interval follows directly from the design. Full-weight correction is harmful for weakly correlated judges (empirical SD ratios up to 1.8 versus human-only), while a tuned weight $\hat\lambda=\widehat{\mathrm{Cov}}_L(y,f)/\mathrm{Var}(f)$ recovers at most small gains: SD ratios 0.96--1.02 across predictors, budgets, and estimands, with coverage 0.92--0.96. Prediction assistance is therefore not worthless in this corpus, only small, and label-fit recalibration adds no consistent precision on top of it.

\begin{table}[H]
\centering
\small
\setlength{\tabcolsep}{5pt}
\caption{Task-grouped 5-fold cross-validated isotonic calibration, ten highest-$R^2$ conditions (worker SD: 1.94). Calibration removes location error. The remaining signal explains at most 8.5\% of individual-rating variance.}
\label{tab:calibration}
\begin{tabular}{lrrrr}
\toprule
Condition & Raw bias & Post-cal.\ bias & CV $R^2$ & Residual SD \\
\midrule
Qwen-FT-2ep 10ex & $-0.67$ & $+0.002$ & 0.085 & 1.85 \\
Qwen-FT-2ep fs+lw & $-0.65$ & $-0.001$ & 0.082 & 1.86 \\
Qwen-FT-1ep fs+lw & $-0.59$ & $-0.001$ & 0.078 & 1.86 \\
EnsHQF average (H+Q+FT) & $-1.19$ & $+0.002$ & 0.077 & 1.86 \\
EnsHQF majority (H+Q+FT) & $-1.18$ & $+0.001$ & 0.064 & 1.87 \\
Qwen-FT-2ep 10ex noThink/t0.7 & $-0.74$ & $+0.001$ & 0.061 & 1.88 \\
Qwen-FT-2ep 5ex think/t1 & $-0.76$ & $-0.001$ & 0.058 & 1.92 \\
Qwen-FT-2ep analogical & $-0.52$ & 0.000 & 0.058 & 1.88 \\
Qwen-FT-2ep 5ex noThink/t0.7 & $-0.75$ & $+0.002$ & 0.057 & 1.88 \\
Qwen-FT-2ep pointwise & $+1.21$ & $+0.002$ & 0.056 & 1.88 \\
\bottomrule
\end{tabular}

\end{table}

\begin{figure}[H]
\centering
\includegraphics[width=\textwidth]{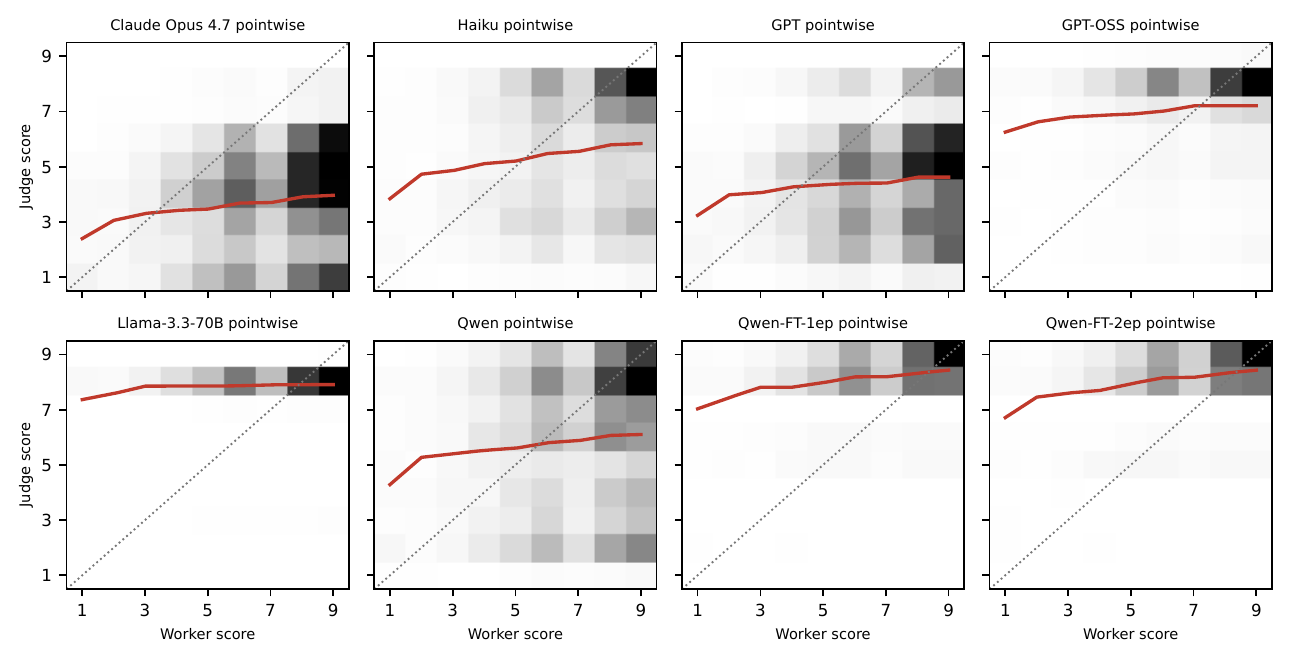}
\caption{Joint distributions of judge and worker scores (grey, share of rows) for eight pointwise configurations on the test split. Red curves are isotonic fits predicting judge score from worker score; the dotted line marks equal scores. The curves describe how judge scores vary across worker ratings. They are distinct from the cross-validated calibrators in Table~\ref{tab:calibration}, which predict worker ratings from judge scores, and should not be interpreted as those calibrators or as a direct explanation of their $R^2$.}
\label{fig:appcalmaps}
\end{figure}

\begin{table}[H]
\centering
\small
\setlength{\tabcolsep}{5pt}
\caption{Simulation of task-level cross-fitted prediction-assisted estimation with isotonic calibration, using 500 repetitions per condition. Entries are interval-width ratios relative to human-only estimation; values above 1 indicate wider intervals. The width-based effective-sample-size summary at the 25\% label budget is $(w_{\mathrm{human}}/w)^2-1$. The final columns report maximum absolute bias and the empirical coverage range of nominal 95\% intervals across budgets; human-only coverage is 0.90--0.97. The interval construction includes the heuristic calibrator-fitting term described in Appendix~\ref{app:calibration}, so the reported coverage assessment applies to this simulation. Across conditions, empirical estimator SD ratios have median 1.00 and range 0.96--1.08. Judge-only estimates retain biases of $+1.21$/$-0.65$/$-1.41$ rating points and $+0.29$/$-0.10$/$-0.17$ in acceptance share.}
\label{tab:ppi}
\begin{tabular}{lrrrrrrr}
\toprule
Predictor & 5\% & 10\% & 25\% & 50\% & Eff.\ $n$ gain at 25\% & $|$Bias$|_{\max}$ & Coverage \\
\midrule
\multicolumn{8}{l}{\emph{Mean rating}} \\
\quad FT2 pointwise & 1.064 & 1.028 & 1.018 & 1.034 & -3.4\% & 0.006 & 0.95--0.96 \\
\quad FT2 listwise & 1.078 & 1.052 & 1.042 & 1.052 & -7.9\% & 0.004 & 0.95--0.97 \\
\quad Haiku pointwise & 1.120 & 1.079 & 1.065 & 1.072 & -11.8\% & 0.013 & 0.94--0.98 \\
\quad TF--IDF (response only) & 1.089 & 1.055 & 1.035 & 1.045 & -6.6\% & 0.009 & 0.94--0.96 \\
\addlinespace
\multicolumn{8}{l}{\emph{Acceptance (score $\geq7$)}} \\
\quad FT2 pointwise & 1.067 & 1.039 & 1.029 & 1.043 & -5.6\% & 0.001 & 0.94--0.98 \\
\quad FT2 listwise & 1.084 & 1.055 & 1.046 & 1.063 & -8.6\% & 0.000 & 0.95--0.97 \\
\quad Haiku pointwise & 1.109 & 1.076 & 1.062 & 1.074 & -11.4\% & 0.002 & 0.92--0.97 \\
\quad TF--IDF (response only) & 1.084 & 1.052 & 1.035 & 1.049 & -6.6\% & 0.004 & 0.95--0.96 \\
\addlinespace
\bottomrule
\end{tabular}

\end{table}

\begin{table}[H]
\centering
\small
\setlength{\tabcolsep}{5pt}
\caption{Calibrator--estimand match under task-grouped 5-fold cross-validation, summarized over the 32 near-complete configurations (median and maximum absolute error). Mean-targeted calibrators recover the mean rating almost exactly, but thresholding their output does not recover the acceptance share; calibrating the acceptance probability directly does. The fold-mean constant shows that near-zero pooled error is attainable with no judge input, so aggregate unbiasedness alone is not evidence of judge information.}
\label{tab:caltargets}
\begin{tabular}{lrrrr}
\toprule
 & \multicolumn{2}{c}{$|$Acceptance err.$|$ (pct.\ pts)} & \multicolumn{2}{c}{$|$Mean err.$|$ (rating pts)} \\
\cmidrule(lr){2-3}\cmidrule(lr){4-5}
Calibrator & median & max & median & max \\
\midrule
Fold-mean constant (no judge input) & 0.0 & 0.0 & 0.001 & 0.001 \\
Raw score, threshold at 7 & 24.0 & 58.1 & 1.223 & 3.231 \\
Intercept-corrected, threshold at 7 & 11.2 & 35.2 & 0.002 & 0.003 \\
Affine-calibrated, threshold at 7 & 11.6 & 39.6 & 0.001 & 0.002 \\
Isotonic mean-calibrated, threshold at 7 & 10.7 & 39.6 & 0.001 & 0.003 \\
Direct binary isotonic (mean probability) & 0.0 & 0.1 & -- & -- \\
\bottomrule
\end{tabular}

\end{table}

\begin{table}[H]
\centering
\small
\setlength{\tabcolsep}{5pt}
\caption{Tuned-$\lambda$ heuristic (cluster-sum plug-in weight) under the same cross-fitted design: CI-width ratio ranges versus the human-only estimator over label fractions 0.05--0.50. Tuning does not produce consistent gains here either (empirical SD ratio median 1.00), and the LLM judges are no better predictors than the response-only TF--IDF control.}
\label{tab:ppiextras}
\begin{tabular}{lrrrr}
\toprule
 & \multicolumn{2}{c}{Mean rating} & \multicolumn{2}{c}{Acceptance (score $\geq7$)} \\
\cmidrule(lr){2-3}\cmidrule(lr){4-5}
Predictor & cross-fit & tuned & cross-fit & tuned \\
\midrule
FT2 pointwise & 1.018--1.064 & 1.014--1.054 & 1.029--1.067 & 1.019--1.038 \\
FT2 listwise & 1.042--1.078 & 1.038--1.070 & 1.046--1.084 & 1.035--1.056 \\
Haiku pointwise & 1.065--1.120 & 1.060--1.104 & 1.062--1.109 & 1.047--1.058 \\
TF--IDF (response only) & 1.035--1.089 & 1.029--1.073 & 1.035--1.084 & 1.019--1.033 \\
\bottomrule
\end{tabular}

\end{table}

\begin{figure}[H]
\centering
\includegraphics[width=\textwidth]{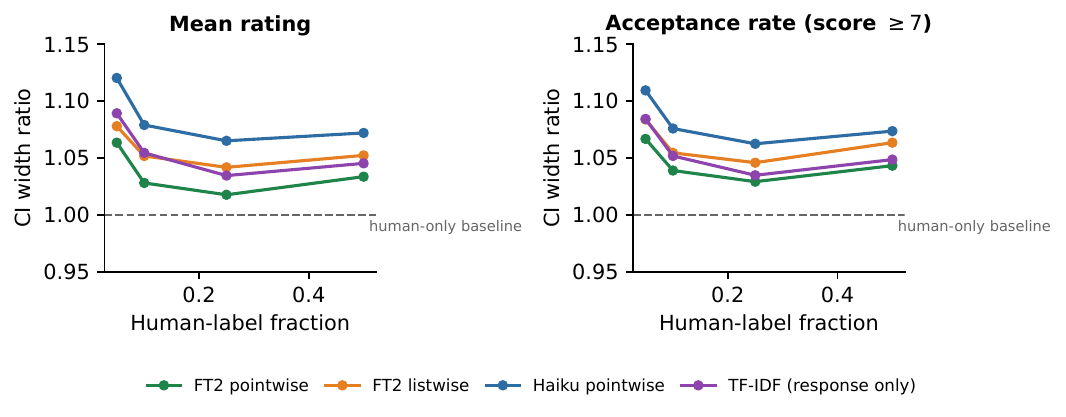}
\caption{CI-width ratio of the cross-fitted prediction-assisted estimator relative to the human-only estimator, by labeled fraction, for the four predictors of Table~\ref{tab:ppi}; values above one are wider.}
\label{fig:ppi}
\end{figure}

\section{Occupation classification: intervals and sensitivity}
\label{app:occ}

\begin{table}[H]
\centering
\small
\setlength{\tabcolsep}{5pt}
\caption{Occupation classification at scores $\geq7$ (strict majority; 307 occupations with $\geq5$ ratings; workers: 0.687 majority-acceptable). Intervals from 1{,}000 occupation-cell bootstrap draws. Each reported configuration disagrees with the sampled worker-majority classification for at least 30.9\% of occupations; the final row is a constant classifier that marks every occupation acceptable.}
\label{tab:occci}
\begin{tabular}{lrr}
\toprule
Judge & Majority-acceptable share & Disagreement with workers \\
\midrule
Claude Opus 4.7 (pointwise) & 0.003 {\scriptsize$[0.000,0.010]$} & 0.684 {\scriptsize$[0.635,0.736]$} \\
Claude Haiku 4.5 (pointwise) & 0.378 {\scriptsize$[0.326,0.430]$} & 0.505 {\scriptsize$[0.450,0.564]$} \\
Qwen-FT-2ep (pointwise) & 0.974 {\scriptsize$[0.954,0.990]$} & 0.313 {\scriptsize$[0.264,0.362]$} \\
Qwen-FT-2ep (few-shot/listwise) & 0.479 {\scriptsize$[0.423,0.534]$} & 0.489 {\scriptsize$[0.436,0.541]$} \\
GPT-OSS-120B (pointwise) & 0.873 {\scriptsize$[0.834,0.909]$} & 0.309 {\scriptsize$[0.261,0.365]$} \\
TF--IDF full context (control) & 0.410 {\scriptsize$[0.358,0.469]$} & 0.466 {\scriptsize$[0.407,0.521]$} \\
\midrule
Constant ``acceptable'' baseline & 1.000 & 0.313 \\
\bottomrule
\end{tabular}

\end{table}

Sensitivity: the disagreement pattern is stable across thresholds $\tau\in\{5,6,7,8\}$ and minimum cell sizes $\{3,5,8\}$. For Opus, disagreement is 69.4\% at $\tau=5$, 88.3\% at $\tau=6$, 68.4\% at $\tau=7$, and 48.2\% at $\tau=8$, largest where judge and worker acceptance curves are steepest. Discrimination is uniformly weak: balanced accuracy against the sampled worker majority is 0.50--0.58 across the six reported instruments, versus 0.50 for the constant baseline, so disagreement rates near one third do not indicate occupational discrimination.

\paragraph{Stability of the worker reference.} Occupation classifications rest on sampled majorities, so we flag an occupation as stable when a within-occupation bootstrap 95\% interval for the worker acceptable share excludes one half, resampling task instances rather than individual ratings (1{,}000 draws; the median occupation has two instances, so certification is coarse and effectively requires agreement across clusters). Of the 307 occupations, 102 qualify; a rating-level bootstrap gives 70, and a margin rule (share $\leq0.40$ or $\geq0.60$) retains 256, with the same pattern under all three rules. On the 102 instance-stable occupations, disagreement across the 32 near-complete configurations spans 14.7\% to 84.3\% with a median of 65.7\% (margin rule: 22.7\% to 76.6\%, median 59.2\%), against 30.6\% to 68.4\% on all 307. The stable set is not a neutral sample: 86 of the 102 (84\%) are majority-acceptable for workers, so harsh judges disagree more often there (Opus rises from 68.4\% to 83.3\%) while the lenient pointwise fine-tune falls from 31.3\% to 16.7\%. Classification disagreement is therefore not an artifact of occupations near the 50\% boundary, but stable-subset rates must be read against this composition.

\paragraph{Aggregation weighting.} The acceptance estimand is insensitive to the aggregation weights in this corpus: the workers' share is 61.1\% response-weighted, 61.5\% with equal task weights, and 61.6\% with equal occupation weights, and the corresponding within-judge shifts have median magnitude below half a point, leaving the 3.0\%--97.9\% cross-judge range essentially unchanged. Because every configuration scores the same responses, the task and generator mixture is held fixed across instruments by design.

\section{Hybrid estimator across matched pairs}
\label{app:hybrid}

Within each instance, the hybrid assigns the sorted pointwise values according to the listwise ordering. Every task, occupation, and family mean is unchanged by construction, so aggregate correlations equal pointwise exactly. It requires both judgments per instance and inherits pointwise location, including its bias.

\begin{table}[H]
\centering
\small
\setlength{\tabcolsep}{4pt}
\caption{Response Spearman for the hybrid across all six matched pairs; boldface marks estimates whose intervals exclude zero. ``Recovered'' is the share of each pair's listwise local gain.}
\label{tab:hybrid}
\begin{tabular}{lrrrrr}
\toprule
Pair & Pointwise & Listwise & Hybrid & $\Delta$ vs.\ pointwise [95\% CI] & Recovered \\
\midrule
Qwen base & .109 & .219 & .150 & \textbf{$+$.041} [$+$.027,$+$.055] & 37\% \\
Haiku & .149 & .219 & .162 & \textbf{$+$.013} [$+$.004,$+$.021] & 18\% \\
GPT-OSS & .093 & .203 & .153 & \textbf{$+$.059} [$+$.038,$+$.081] & 54\% \\
GPT & .091 & .151 & .100 & $+$.010 [$-$.000,$+$.020] & 16\% \\
Qwen FT1 & .215 & .258 & .219 & $+$.004 [$-$.009,$+$.015] & 9\% \\
Qwen FT2 & .214 & .263 & .228 & \textbf{$+$.013} [$+$.001,$+$.027] & 27\% \\
\bottomrule
\end{tabular}
\end{table}

\section{Field-specific agreement}
\label{app:fields}

The pointwise FT2 judge aligns more with holistic workforce judgments than with individual rubric criteria, cautioning against treating ``judge quality'' as one latent capability.

\begin{table}[H]
\centering
\small
\caption{Per-field agreement for the pointwise FT2 judge (``accuracy not applicable'' treated as missing).}
\label{tab:fields}
\begin{tabular}{lrrrr}
\toprule
Worker target & $n$ & Response $\rho$ & $\pacc$ & Pair coverage \\
\midrule
Understandability & 4,501 & .122 & .537 & .249 \\
Relevance & 4,501 & .127 & .541 & .187 \\
Factual accuracy & 4,241 & .097 & .531 & .182 \\
Tone & 4,501 & .114 & .541 & .181 \\
Employee quality & 4,501 & .236 & .621 & .529 \\
Manager acceptance & 4,501 & .214 & .596 & .510 \\
\bottomrule
\end{tabular}
\end{table}

\begin{figure}[H]
\centering
\includegraphics[width=0.95\textwidth]{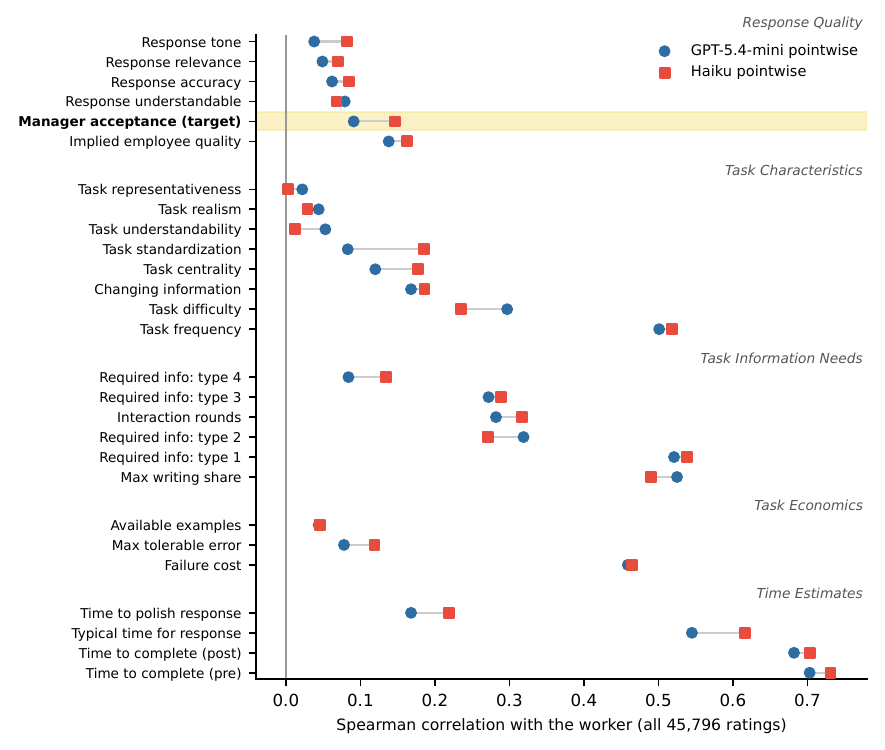}
\caption{Per-field Spearman correlation with the worker over the full corpus (45{,}796 ratings) for the two full-corpus pointwise runs. Time-to-complete estimates, task frequency, the primary required-information type, and failure cost reach $\rho=0.46$--$0.73$; every graded response-quality field, including the target (highlighted), stays below $0.17$ for both judges.}
\label{fig:appfields}
\end{figure}

We additionally evaluate GPT-5.4-mini and Haiku pointwise over the full 45{,}796-rating corpus. Neither configuration was fine-tuned on the corpus's worker labels, and both prediction sets were generated in May 2026, before the audit was designed. These full-corpus results provide a descriptive comparison across survey fields, rather than a new task-disjoint confirmation of analyses developed on the test split. Agreement is higher for several task characteristics and time estimates than for response-quality ratings (Figure~\ref{fig:appfields}): every graded response-quality field, including \texttt{manager\_accept}, has Spearman correlation at most 0.17. Thus, agreement varies substantially across fields. Differences in field scales, response distributions, and reference reliability limit interpretation of this pattern as a difference in underlying judging capability.

\section{Reliability of the human cell means}
\label{app:reliability}

Across 200 random splits of raters within cells, the median correlation between occupation-level half-sample means is 0.049, corresponding to a Spearman--Brown estimate of 0.093 over 237 cells \citep{brown1910spearman}. Splitting requires at least two distinct raters per cell, which retains 237 of the 307 evaluable occupations and 380 of the 515 task IDs. The task- and family-level Spearman--Brown estimates are 0.065 and 0.731. If the occupation estimate is interpreted as classical reliability, and the relevant measurement-error assumptions hold, attenuation theory gives an approximate maximum correlation of $\sqrt{0.093}\approx0.30$ with an error-free latent occupation score \citep{spearman1904attenuation}. This is assumption-dependent context, not a universal ceiling for every judge--worker correlation. The reliability estimate is also uncertain: the reported interval on the uncorrected split-half correlation is 0.010--0.087, and an independent rater split yields corrected reliability 0.100, corresponding to $\sqrt{0.100}\approx0.316$. These calculations combine variation across raters with task sampling within occupations and cannot replace repeated ratings of the same response. Independent mean-zero target noise contributes no expected mean bias, but finite-sample variation and nonlinear thresholding remain relevant to the audit's mean, acceptance-share, and classification estimates.

\section{Fine-tuning configuration}
\label{app:ft}

The Qwen3.5-9B LoRA run uses the 36{,}649-row training split and the 4{,}646-row validation split for loss monitoring only. LoRA rank 16, $\alpha=32$, dropout 0.05, adapters on all linear modules; AdamW, learning rate $5\times10^{-5}$, cosine decay, warmup ratio 0.03, effective batch size 8, sequence packing, max gradient norm 1, seed 42. Checkpoints retained after one and two epochs; validation loss 0.0726 $\to$ 0.0721. Fine-tuning used the original pointwise full-schema prompt format, so the listwise arms evaluate the checkpoint under a transferred prompt format, consistent with reading the reversal as a bundled protocol contrast (\S\ref{sec:reversal}). At matched settings, the intervals for FT2-minus-base and FT2-minus-FT1 differences in response Spearman correlation and pair accuracy include zero. These comparisons do not establish an ordering improvement from fine-tuning, nor do they establish equivalence. The pointwise score distributions show a substantial change in location, which we describe separately below.

\paragraph{Score concentration under greedy decoding.} Training labels have mean 6.92, with 49.7\% at scores 8--9 and a mode of 9 (27\%). Under temperature-0 decoding, the two-epoch pointwise checkpoint assigns 8 or 9 to 89.8\% of test responses, never assigns 7, and has mean score 8.18. This is 1.26 rating points above the training-label mean and 1.21 rating points above the test-worker mean of 6.97. The concentration near the upper end of the scale accounts for the high acceptance share in \S\ref{sec:severity}. It is compatible with mode-seeking output under greedy decoding, but these observations do not isolate decoding from the effects of fine-tuning, prompting, or the label distribution.

\section{Serving-stack reproduction check}
\label{app:serving}

The fine-tuned checkpoint requires hybrid reasoning to be disabled through the serving stack's chat-template flag. Some deployments accept an alternative request field and ignore it silently. In a validation-split attempt, listwise parse coverage fell to 22.7\% (the model emitted one rating object instead of a list) and pointwise scores shifted without any request error; that attempt was discarded and rerun with the corrected flag. The partial-coverage reasoning variant in the registry (22.3\% of test rows) is a separate occurrence with the same signature, retained as the audited instance; both trace to the ignored thinking-disable flag. Reproductions should compare a small sample against the released predictions before evaluating a full split.

\end{document}